\documentclass[a4paper,fleqn]{cas-sc}

\usepackage[authoryear]{natbib}
\hypersetup{ colorlinks, citecolor=blue, linkcolor=blue, urlcolor=Blue}
\def\tsc#1{\csdef{#1}{\textsc{\lowercase{#1}}\xspace}}
\tsc{WGM}
\tsc{QE}
\tsc{EP}
\tsc{PMS}
\tsc{BEC}
\tsc{DE}

\usepackage{upgreek}

\usepackage{color,soul}
\newcommand{\highlighting}[1]{%
    {%
    \sethlcolor{white}%
    \hl{#1}%
    }%
}

\begin{document}
\let\WriteBookmarks\relax
\def\floatpagepagefraction{1}
\def\textpagefraction{.001}

\shorttitle{MixerCA: An Efficient Model for HSI Classification}

\shortauthors{M.Q. Alkhatib, A. Jamali}

\title [mode = title]{MixerCA: An Efficient and Accurate Model for High-Performance Hyperspectral Image Classification}

\author[1]{Mohammed Q. Alkhatib}[orcid=0000-0003-4812-614X]
\cormark[1]
\credit{Conceptualization, Methodology, Software, Data curation, Writing - Original draft preparation}
\ead{mqalkhatib@ieee.org}

\author[2]{Ali Jamali}[orcid=0000-0002-6073-5493]
\cormark[1]
\credit{Conceptualization, Methodology, Software, Data curation, Writing - Original draft preparation}
\ead{alij@sfu.ca}

\affiliation[1]{organization={College of Engineering and IT, University of Dubai},
            city={Dubai},
            postcode={14143}, 
            country={UAE}}

\affiliation[2]{organization={Department of Geography, Simon Fraser University},
            addressline={8888 University Dr}, 
            city={Burnaby},
            postcode={BC V5A 1S6}, 
            country={Canada}}

\begin{abstract}
Over the past decade, hyperspectral image (HSI) classification has drawn considerable interest due to HSIs' ability to effectively distinguish terrestrial objects by capturing detailed, continuous spectral information. The strong performance of recent deep learning techniques in tasks like image classification and semantic segmentation has led to their growing use in HSI classification, {due to their ability to capture} complex spatial and spectral features more effectively than traditional methods. This paper presents MixerCA, a novel lightweight model for HSI classification that leverages depthwise convolution and a self-attention mechanism. MixerCA {integrates depth-wise convolutions, token and channel mixing, and coordinate attention into a unified structure} to decouple spatial and channel interactions, {maintain consistent resolution throughout the network}, and directly process HSI patches. Extensive experiments on four hyperspectral benchmark datasets reveal MixerCA’s clear advantages over several competing algorithms, including 2D-CNN, 3D-CNN, Tri-CNN, HybridSN, ViT, and Swin Transformer. {The source code is publicly available at} \url{https://github.com/mqalkhatib/MixerCA}.

\end{abstract}



\maketitle

\section{Introduction}
\label{sec:introduction}
Hyperspectral Imaging (HSI) systems capture spectral and spatial information concurrently across hundreds of narrow, contiguous spectral bands, spanning wavelengths from the visible to the infrared spectrum. HSI has a wide array of applications in remote sensing, including environmental monitoring \citep{stuart2019hyperspectral, stuart2022high}, agriculture \citep{barbedo2023review, lu2020recent}, mineral exploration \citep{hajaj2024review, okada2024advanced}, and surveillance \citep{kutuk2023ground, gao2024appearance}. Each pixel in an HSI image encodes a spectrum of hundreds of values, representing the reflected or emitted energy across different wavelengths—referred to as the spectral response. Unlike traditional images, HSIs provide detailed spectral data, revealing the chemical and physical properties of materials. This rich spectral information, based on the interaction of light with atomic and molecular structures, significantly enhances the efficacy of material classification tasks.

HSI classification refers to the process of assigning each pixel in an image to a specific class, facilitating data interpretation by leveraging the spectral profile of each pixel to reveal material composition. Although machine learning classifiers such as SVM \citep{melgani2002support} and Random Forests \citep{joelsson2005random, wang2016airborne} have advanced, they frequently encounter difficulties in capturing the complex, non-linear spectral correlations present in HSIs. The high-dimensional dependencies among spectral bands present challenges for SVMs without computationally intensive non-linear kernels, and Random Forests often fail to capture these subtle interdependencies accurately. Additionally, these methods rely solely on spectral data, disregarding spatial context, which limits classification performance.

Recent advances in deep learning, especially through Convolutional Neural Networks (CNNs), have greatly enhanced classification accuracy by automating hierarchical feature extraction \citep{makantasis2015deep, chen2016deep}. CNNs, which have gained significant attention in computer vision, are designed to learn and extract features from raw data through layered hierarchies: shallower layers capture basic edges and textures, while deeper layers identify more complex patterns. This capability has led to the widespread adoption of CNNs in HSI classification, where they are effectively utilized to extract rich spectral-spatial information for improved performance \citep{makantasis2015deep, chen2016deep}. The authors in \citep{gao2018hyperspectral} and \citep{vaddi2020hyperspectral} have demonstrated the efficacy of CNNs in HSI classification, achieving notable improvements in classification accuracy. In \citep{gao2018hyperspectral}, a multi-feature learning framework is introduced, combining spectral and spatial features within a CNN to enhance classification accuracy. Similarly, \citep{vaddi2020hyperspectral} proposed a framework that combines data normalization, Probabilistic Principal Component Analysis (PPCA), and Gabor filtering to integrate spectral and spatial information for robust classification. These studies underscore the flexibility of CNNs in handling diverse HSI applications through spectral-spatial feature fusion.

Further developments in hybrid and advanced CNN architectures for HSI classification have been explored by the authors in \citep{roy2019hybridsn}, \citep{medus2021hyperspectral}, \citep{alkhatib2023tri}, and \citep{zhong2022hyperspectral}. In \citep{roy2019hybridsn}, the HybridSN model is proposed, incorporating both 3D and 2D CNN layers to efficiently capture spatial-spectral information. Likewise, \citep{medus2021hyperspectral} explored real-time anomaly detection in food tray packaging using CNNs to achieve high accuracy in detecting contamination in an industrial setting. Later, \citep{alkhatib2023tri} introduced the Tri-CNN model, a multi-branch CNN utilizing 3D CNN layers at different scales for improved spectral-spatial feature extraction, yielding superior results across various datasets. Finally, \citep{zhong2022hyperspectral} developed the PMI-CNN, a multi-input, parallel architecture that employs separate convolutional branches to capture spectral-spatial dependencies, resulting in significant classification accuracy gains on benchmark datasets. Collectively, these studies highlight the effectiveness of CNN-based frameworks in addressing the high dimensionality and complexity of HSI data through spectral-spatial fusion, multi-branch, and hybrid architectures.

Despite their effectiveness, both 2D-CNN and 3D-CNN architectures present limitations when applied to HSI classification. 2D-CNN architectures, while effective at capturing spatial information, struggle to extract highly discriminative feature maps from the spectral dimension. In contrast, 3D-CNNs can capture spectral-spatial features more comprehensively, yet they are computationally intensive due to the large number of 3D convolution operations required. Deep 3D-CNN models also demand substantial training data, which is challenging to meet given the limited sample sizes in most publicly available HSI datasets. Additionally, many 3D-CNN approaches rely on stacking multiple 3D convolutions, which complicates direct optimization of the loss function due to the non-linear structure \citep{yang2020synergistic}.

In order to address the computational demands of traditional CNNs in HSI classification, recent studies have proposed efficient architectures. Gao et al. \citep{gao2020multiscale} and Ye et al. \citep{ye2023computationally} introduce multiscale networks with depthwise separable convolutions to reduce parameter counts and improve feature extraction. Similarly, Dang et al. \citep{dang2020depth} and Cui et al. \citep{cui2021litedepthwisenet} utilize depthwise convolutions with residual structures and pointwise decompositions, achieving lower computational costs. Nguyen et al. \citep{nguyen2024hyperspectral} and Asker et al. \citep{asker2024hybrid} further enhance efficiency with encoder-decoder and hybrid models, incorporating focal loss and 3D depthwise convolutions for improved spatial-spectral feature extraction. Together, these approaches reduce complexity and enhance classification accuracy in HSI tasks.

Attention mechanisms play a critical role in enhancing HSI classification by managing complex data and extracting relevant features. Ma et al. \citep{ma2019double} and Xue et al. \citep{xue2021hresnetam} utilize double-branch and hierarchical architectures to independently extract spectral and spatial features, applying attention mechanisms to improve feature discrimination. Similarly, Shi et al. \citep{shi2022hyperspectral} and Wang et al. \citep{wang2023hyperspectral} employ multiscale and channel spatial attention modules to adaptively weight spectral and spatial features, yielding notable gains in classification accuracy. Liao et al. \citep{liao2024hyperspectral} adopt a vision transformer with multihead attention to capture long-range spatial and spectral dependencies, reducing model parameters while enhancing performance. Additionally, Gunduz et al. \citep{gunduz2024hyperspectral} combine CNNs with gated recurrent units and dual attention to balance spatial and spectral modeling, while Viel et al. \citep{viel2023hyperspectral} integrate attention with transformers, 1D-CNNs, and LSTMs to optimize computational efficiency. Collectively, these studies highlight the effectiveness of attention mechanisms in simplifying model complexity and improving classification accuracy in HSI tasks. A key challenge, however, is managing high-dimensional spectral-spatial data with limited labeled samples, which increases computational demands. This issue can be alleviated by incorporating dimensionality reduction techniques, enabling efficient processing while preserving essential features for accurate classification.

{Unlike prior models that use these mechanisms in isolation, this paper proposes a new model, namely MixerCA, which integrates depth-wise convolutions, token/channel mixing, and coordinate attention into a unified framework optimized for hyperspectral image (HSI) classification. This architecture is specifically designed to extract spatial-spectral features across multiple scales while preserving positional awareness and minimizing computational cost. In contrast to generic lightweight models such as MobileNet or MLP-Mixer, MixerCA directly addresses the spectral complexity and spatial granularity inherent in HSI data, offering a domain-adapted solution that balances accuracy and efficiency.

}

The primary contributions of this article are as follows:
\begin{enumerate}
    \item Development of a lightweight model to enhance efficiency in hyperspectral classification.
    \item By using depth-wise and point-wise convolutions, the developed model reduces the number of parameters and computations compared to traditional convolutional networks, making it more efficient and faster to train and deploy.
    \item The model can capture complex, high-level features across various channels by combining token mixing (for spatial relations) and channel mixing (for feature interaction). This enhances performance on tasks requiring for the learning of intricate patterns.
    \item Comprehensive evaluation of different attention mechanisms to identify the most effective technique for HSI classification.
\end{enumerate}

The structure of this paper is as follows: Section \ref{sec:relatedWork} provides a review of the related work. Section \ref{sec:model} details the proposed MixerCA model. Section \ref{sec:EXPERIMENT} presents the experimental setup and results. Finally, Section \ref{sec:conclusion} discusses the conclusions and outlines directions for future research.

\section{related work} \label{sec:relatedWork}
\subsection{Convolution Overview}
Convolutional layers are essential components in deep learning architectures, especially for processing image data. In a conventional convolution operation, each filter interacts with all input channels simultaneously, creating a combined output that captures inter-channel dependencies. For an input tensor $\mathbf{X}$ with dimensions \((H, W, C_{in})\) (height, width, and input channels), and \(C_{out}\) output channels, this operation can be described as:

\small
\begin{align}
\label{eq:2DConv}
\mathbf{Y}(h, w, c_{out}) = &\sum_{c_{in}} \sum_{i=0}^{K-1} \sum_{j=0}^{K-1} \mathbf{X}(h+i, w+j, c_{in}) \cdot \\ \nonumber
&\mathbf{W}(i, j, c_{in}, c_{out}) + b_{c_{out}}
\end{align}
\normalsize

where $\mathbf{X}$ is the input tensor of size $(H, W, C_{in})$, $\mathbf{W}$ is the weight tensor with dimensions $(K, K, C_{in}, C_{out})$, $b_{c_{out}}$ represents the bias for each output channel, and $\mathbf{Y}$ is the resulting output tensor of size $(H', W', C_{out})$. Each output channel in conventional convolution is computed by applying a filter across all input channels, which combines information across channels but requires high computational resources and numerous parameters.

In contrast, depth-wise convolution reduces computational complexity and parameter count by processing each input channel separately. Instead of combining information across channels, depth-wise convolution applies an independent filter to each channel, avoiding inter-channel mixing. For an input tensor $\mathbf{X}$ of size $(H, W, C_{in})$, depth-wise convolution is defined as:

\small
\begin{align}
\label{eq:DWConv}
\mathbf{Y}(h, w, c_{in}) = &\sum_{i=0}^{K-1} \sum_{j=0}^{K-1} \mathbf{X}(h+i, w+j, c_{in}) \cdot \\ \nonumber
&\mathbf{W}(i, j, c_{in}) + b_{c_{in}}
\end{align}
\normalsize

where $\mathbf{W}$ is now a filter tensor of size $(K, K, 1, C_{in})$, with a unique filter for each input channel, and $b_{c_{in}}$ is the bias for each channel. The output tensor $\mathbf{Y}$ typically retains the same number of channels as the input, with dimensions $(H', W', C_{in})$. Depth-wise convolution thus reduces both parameters and computation, making it ideal for resource-limited applications.

Depth-wise convolutions are used effectively in models like MobileNets \citep{howard2017mobilenet}, where they are combined with point-wise ($1\times1$) convolutions to aggregate the output channels, resulting in lightweight architectures with competitive performance. The main differences between conventional and depth-wise convolutions can be summarized as follows:

\begin{itemize}
    \item \textbf{Conventional Convolution}: Each filter spans all input channels, resulting in \(C_{in} \times C_{out}\) filter parameters and capturing comprehensive features across channels.
    \item \textbf{Depth-wise Convolution}: Each channel has its own independent filter, resulting in only \(C_{in} \times 1\) filters, which significantly reduces both parameters and computational cost without sacrificing model accuracy.
\end{itemize}

Overall, depth-wise convolutions enable the creation of fast, lightweight architectures by lowering memory and computational demands, making them highly suitable for efficient model design without major performance loss.

\subsection{Attention Mechanism} \label{sec:attention}
Attention mechanisms have significantly enhanced deep learning model performance in complex tasks like image classification by focusing on critical input data while suppressing irrelevant features. Various attention modules, including CBAM \citep{woo2018cbam}, SE \citep{hu2018senet}, ECA \citep{wang2020eca}, and CA \citep{hou2021coordinate}, each offer specific benefits. CBAM combines channel and spatial attention, improving feature representation through selective emphasis \citep{woo2018cbam}. SE compresses spatial information, selectively activating channels to refine channel-specific features \citep{hu2018senet, li2019selective}. ECA avoids dimensionality reduction, directly capturing local channel interactions, making it efficient for large datasets \citep{wang2020eca}. CA encodes spatial information with long-range dependencies, maintaining spatial context and channel accuracy, which is advantageous in semantic segmentation \citep{hou2021coordinate}.

In HSI classification tasks, attention mechanisms enhance both spatial and spectral feature extraction. Transformer-based approaches, such as Liao et al.'s vision transformers, capture long-range dependencies without convolutions, reducing parameters \citep{liao2024hyperspectral}. Shu et al. integrate global and local features in a dual attention transformer network (DATN) \citep{shu2024dual}, and Zhao et al. use a multi-attention transformer with adaptive superpixel segmentation to enable efficient training with limited labeled samples \citep{zhao2023hyperspectral}. CNN-based methods, such as Hang et al., employ spectral and spatial attention mechanisms, outperforming standard CNNs by focusing on more discriminative channels or positions \citep{hang2020hyperspectral}.

These studies underscore the efficacy of attention-augmented transformer and CNN architectures in improving classification performance across datasets, with each attention module tailored to enhance feature fidelity, spatial accuracy, or training efficiency.

\section{METHODOLOGY} \label{sec:model}
\subsection{Architecture of the proposed model}
\begin{figure*}[!bt]
\centering
\includegraphics[width=0.85\linewidth]{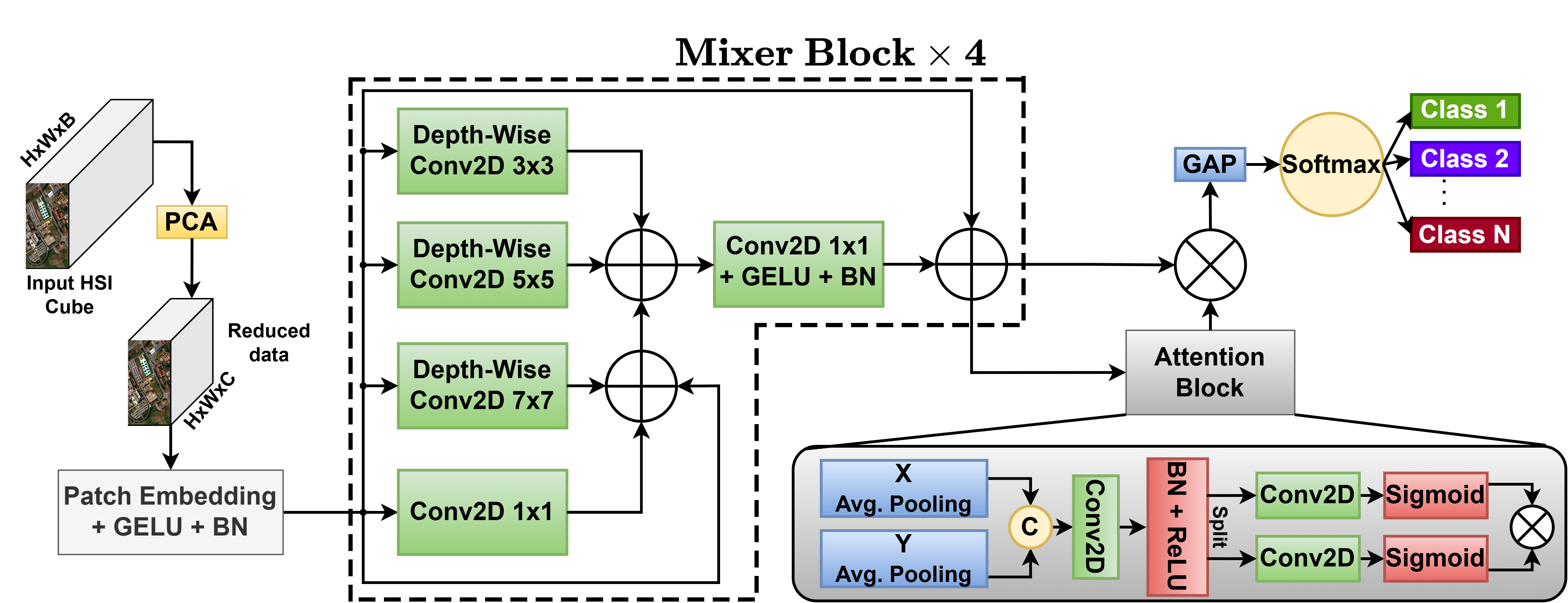}
\caption{The overall architecture of the developed MixerCA deep learning model.}
 \label{fig:model}
\end{figure*}

Figure \ref{fig:model} provides an overview of the proposed model architecture. The MixerCA framework includes several depth-wise 2D convolution layers that capture features at different scales, this process is repeated four times, along with an attention block that brings focus to the most important features. Finally, fully connected layers to perform the classification task. Further details of the proposed framework are presented in the following sections.
{While depth-wise convolutions and token/channel mixing have been explored individually in various lightweight models (e.g., MobileNet, MLP-Mixer), their combined use alongside coordinate attention—specifically adapted for hyperspectral inputs—is what distinguishes MixerCA. The model utilizes parallel multi-scale convolutional paths, spatially-aware attention, and spectral dimensionality reduction to achieve a highly efficient yet accurate architecture suited for pixel-level HSI classification.}

\subsection{Data Preprocessing}
Dimensionality reduction is essential for processing HSI data with deep learning models. In this study, Principal Component Analysis (PCA) \citep{ali2019analysis} is employed for this purpose. PCA is a widely adopted unsupervised technique for reducing dimensionality and extracting relevant features. {This method takes advantage of the strong correlation among adjacent HSI bands, which often contain redundant information about the observed materials.} PCA operates through an orthogonal transformation, which takes a set of potentially correlated variables and projects them onto a set of linearly uncorrelated values known as Principal Components (PCs). Given an HSI cube of dimensions $H \times W \times B$—where $H$ and $W$ represent the spatial dimensions and $B$ is the number of spectral bands—PCA reduces the depth of the cube from $B$ to $C$, resulting in a transformed HSI cube of size $H \times W \times C$, where $C$ is the new spectral dimension and $C \ll B$.

\subsection{Mixer Module }
In the MixerCA model, token mixing and channel mixing are key operations that enable the model to capture both spatial and feature interactions, similar to MLP-Mixer \citep{Tolsti2021} and SGU-MLP \citep{SGUMLP}. Token mixing is performed by applying a fully connected layer (MLP) to mix information across spatial patches of the input, {enabling the model to learn long-range spatial dependencies.} Channel mixing, on the other hand, is achieved through point-wise convolutions ($1\times1$ convolutions), which mix features across different channels while maintaining the spatial dimensions. {Together, these operations allow MixerCA to learn complex spatial-spectral patterns more effectively}. 
{Let the input image be represented as} \( X \in \mathbb{R}^{H \times W \times C} \), where \( H \) and \( W \) denote the spatial dimensions (height and width), and \( C \) is the number of spectral channels. {The notation used throughout this section is introduced where needed to ensure clarity.} \( X \) represents the input tensor, while \( X_i \) and \( Z_i \) denote intermediate inputs and outputs in different blocks. \( K_i \) and \( F_i \) refer to the kernel size and number of filters in the \( i^\text{th} \) {layer.}

In the proposed architecture, the input image tensor is initially passed through a convolutional layer (eq. \ref{eq:2DConv}) with a kernel $K_1$, a stride size of $1\times1$, expressed as:

\begin{equation}
Z_1 = \text{Conv2D}(X, K_1, F_1)
\end{equation}

where \( K_1 \) is the kernel size (set to 1), and \( F_1 \) is the number of filters. This layer performs a basic transformation on the input image tensor, producing feature maps with increased depth while maintaining the spatial dimensions of the input. {The use of a $1\times1$ convolutional layer with 64 filters and a stride of 1 offers several advantages.} {First, it increases the number of feature channels without altering the spatial resolution, allowing the model to learn more abstract and hierarchical feature representations.} {Second, it is computationally efficient compared to larger kernels (e.g., $7\times7$) since it processes each pixel independently, without incorporating information from neighboring pixels.} {Third, the multiple filters support nonlinear feature learning by extracting diverse characteristics from the input image.} Overall, by enhancing the model's ability to recognize intricate patterns while preserving computational efficiency, this layer contributes to improved performance. Next, the feature map \( Z_1 \) undergoes an activation block. The GELU activation function is applied to \( Z_1 \), followed by batch normalization to produce \( Z_2 \).

The output \( Z_2 \) is then passed through the mixer block. {Within each mixer block, the input} \( X_i \) {is processed using multiple depth-wise convolution layers with varying kernel sizes, as defined in equation} \ref{eq:DWConv}:

\begin{equation}
Z_{1i} = \text{DepthwiseConv2D}(X_i, K_3)
\end{equation}

\begin{equation}
Z_{2i} = \text{DepthwiseConv2D}(X_i, K_5)
\end{equation}

\begin{equation}
Z_{3i} = \text{DepthwiseConv2D}(X_i, K_7)
\end{equation}

where \( K_3 \), \( K_5 \), and \( K_7 \) are kernel sizes 3, 5, and 7, respectively. {These kernel sizes were selected to extract spatial information at multiple scales. The 3×3 kernel captures fine local details, while the 5×5 and 7×7 kernels provide broader contextual understanding. This multi-scale strategy enhances the model’s ability to recognize structures of varying sizes within the scene, which is particularly important in hyperspectral image classification tasks.}

Depth-wise convolutions offer several advantages, particularly for large or complex models. By applying each filter to a single input channel—unlike standard convolutions that operate across all channels—they substantially reduce computational complexity. This efficiency is especially valuable when processing hyperspectral data, where the input tensor contains a large number of channels. The reduction in operations and parameters translates to faster computations, making depth-wise convolutions ideal for real-time applications. Moreover, they enable the model to capture fine-grained spatial details independently for each channel, which is beneficial for learning per-channel features. When used with appropriate padding, they also preserve spatial resolution, which is critical in pixel-level tasks. Depth-wise convolutions are key components in lightweight architectures such as MobileNets \citep{howard2017mobilenet}, {which are designed to balance accuracy and efficiency. Their ability to reduce memory usage and processing time while retaining key spatial features makes them particularly suitable for resource-constrained environments.} Additionally, a standard convolution operation is applied to the input image tensor:

\begin{equation}
Z_{4i} = \text{Conv2D}(X_i, F_1, 1)
\end{equation}

The outputs of the depth-wise convolutions and the convolutional operation are aggregated with a residual connection, defined by:

\begin{equation}
X_{i+1} = X_i + Z_{1i} + Z_{2i} + Z_{3i} + Z_{4i}
\end{equation}

Following this, a point-wise convolution ($1\times1$ convolution) is applied:

\begin{equation}
X_{i+1} = \text{Conv2D}(X_{i+1}, F_2, 1)
\end{equation}


where \( F_2 \) represents the number of filters. {Point-wise convolutions ($1\times1$ kernels) offer several benefits. By operating independently on each pixel in every channel, they reduce the number of parameters and enhance model efficiency while preserving spatial dimensions.} {They also enable channel-wise feature integration, allowing the model to learn more expressive and complex representations.} {Due to their low computational cost, point-wise convolutions support faster training and inference, making them suitable for deployment in resource-constrained environments such as mobile devices.} {They are also fundamental components in lightweight architectures like MobileNets} \citep{howard2017mobilenet}, {which are designed for efficiency through modular and parameter-efficient layers.} A final residual connection is added:

\begin{equation}
X_{i+1} = X_{i+1} + X_i
\end{equation}

After processing through all the mixer blocks (i.e., 4 blocks), the output map is sent to  the coordinate attention (CA) module (sec \ref{sec:CA}), expressed as: 

\begin{equation}
\tilde{X} = CA (X_{i+1})
\end{equation}

The output map of the attention module is then passed through a global average pooling, defined by:

\begin{equation}
Z_{\text{avg}} = \frac{1}{H' W'} \sum_{h=1}^{H'} \sum_{w=1}^{W'} \tilde{X}[h, w]
\end{equation}

where \( H' \) and \( W' \) represent the height and width after the final convolutional operation. The pooled feature vector \( Z_{\text{avg}} \) is then passed through a dense layer for the final classification, expressed as:

\begin{equation}
\text{logits} = \text{Dense}(Z_{\text{avg}}, C_{\text{out}}, \text{softmax})
\end{equation}

where \( C_{\text{out}} \) is the number of output classes. The softmax function converts the logits into class probabilities:

\begin{equation}
\hat{y} = \text{softmax}(\text{logits})
\end{equation}

Finally, \( \hat{y} \) represents the predicted class label.

\subsection{Coordinate Attention (CA)}
\label{sec:CA}
{The Coordinate Attention (CA) module is applied after the mixer block to refine the extracted features by capturing long-range dependencies along spatial and channel dimensions with minimal computational overhead. Given the input feature tensor $\mathbf{X}$,which is the output of the preceding mixer block,} average pooling is performed along the height ($H$) and width ($W$) dimensions to encode positional information, resulting in:

\begin{equation}
x_h = \frac{1}{W} \sum_{i=0}^{W-1} \mathbf{X}(H, i, C), 
\end{equation}

\begin{equation}
x_w = \frac{1}{H} \sum_{i=0}^{H-1} \mathbf{X}(i, W, C),
\end{equation}

where $x_h$ and $x_w$ correspond to operations over the height $H$ and width $W$, respectively.

The dimension of $x_h$ is then reshaped to ${1 \times H \times C}$ and concatenated with $x_w$. A shared $1 \times 1$ convolution operation $F_s$ is applied to the concatenated result:

\begin{equation}
l = \psi(F_s([x_h, x_w])),
\end{equation}

where $[ \cdot, \cdot ]$ represents concatenation, $\psi(\cdot)$ denotes a non-linear activation function, specifically \text{ReLU}, and $r$ is the reduction ratio controlling block size.

The output $l$ is then split into $l^h$ and $l^w$ along the spatial dimension. Separate $1 \times 1$ convolution operations, $F_h$ and $F_w$, are applied to $l^h$ and $l^w$, respectively, to match the channel count of $\mathbf{X}$:

\begin{equation}
g^h = \sigma(F_h(l^h)), 
\end{equation}

\begin{equation}
g^w = \sigma(F_w(l^w)),
\end{equation}

where $\sigma(\cdot)$ is the sigmoid activation function. The attention weight $\mathbf{W}$, derived from $\mathbf{X}$, is then given by:

\begin{equation}
\mathbf{W} = g^h \times g^w, 
\end{equation}

Finally, the output of the attention block, $\mathbf{\tilde{X}}$, is:

\begin{equation}
\mathbf{\tilde{X}} = \mathbf{X} \odot \mathbf{W}, 
\end{equation}

where $\odot$ denotes element-wise multiplication.

\subsection{Loss Function}
Cross-Entropy (CE) is used as the loss function in this paper due to its effectiveness in multiclass classification tasks. It measures the discrepancy between the predicted probability distribution and the true labels, ensuring that the predicted probabilities align closely with the reference labels. The mathematical formulation is expressed as:
\begin{equation}
Loss_{CE} = -\frac{1}{M} \sum_{m = 1}^M \sum_{l = 1}^L y_l^m \log(\hat{y}_l^m)
\end{equation}
where $y_l^m$ and $\hat{y}_l^m$ are the reference and predicted labels, respectively, $M$ and $L$ are the overall number of small batch samples and land cover categories, respectively.

This loss function is particularly effective for multiclass classification, especially in imbalanced scenarios, as its logarithmic nature ensures proportional contributions from all classes to the overall loss. Combined with softmax activation at the output layer, Cross-Entropy produces normalized probability distributions, enhancing interpretability and optimization. Its adoption in this study aligns with the goal of achieving accurate classification across diverse land cover categories.

\section{EXPERIMENT AND ANALYSIS}
\label{sec:EXPERIMENT}
This section will provide an overview of the HSI datasets utilized in this study. Next, the experimental setup and parameter analysis will be introduced. Following this, a comparison of various attention mechanisms will be conducted, highlighting the justification for choosing coordinate attention. Lastly, the proposed model will be evaluated against existing methods to demonstrate its superiority and effectiveness through both quantitative and qualitative analysis.

\subsection{Datasets} \label{sec:dataset}
In our experiments, we utilize two of the most widely used HSI datasets: Pavia University and Salinas. Additionally, the Gulfport of Mississippi dataset is included, despite its limited use in HSI classification, due to its small size and composition of only 72 spectral bands, making it of particular interest. We also incorporate the Xuzhou dataset, captured in 2014. Each of these datasets has unique characteristics, detailed as follows:

\begin{enumerate}
    \item Pavia University (PU): This scene was captured by the Reflective Optics Imaging Spectrometer Sensor (ROSIS) during an airborne campaign over Pavia, northern Italy. The image includes 103 spectral bands, covering wavelengths from 0.43 to 0.86 $\upmu$m, with a spatial resolution of 1.3 meters. The dimensions of Pavia University are $610 \times 340$ pixels. Figure~\ref{fig:PU_Results}a presents an RGB color composite of the scene, while the reference classification map in Figure~\ref{fig:PU_Results}b illustrates nine classes, with unassigned pixels shown in black and labeled as Unassigned.
   
    \item Salinas (SA): This scene was captured by the Airborne Visible/Infrared Imaging Spectrometer (AVIRIS) sensor over Salinas Valley, California. The image comprises 224 spectral bands, covering wavelengths from 0.4 to 2.45 $\upmu$m, with a spatial resolution of 3.7 meters. The dimensions of the Salinas image are $512 \times 217$ pixels. Due to distortions from water absorption, bands 108--112, 154--167, and 224 were excluded. Figure~\ref{fig:SA_Results}a displays an RGB composite of the scene, while the reference classification map in Figure~\ref{fig:SA_Results}b indicates that the Salinas ground truth map includes 16 classes.

    \item Mississippi Gulfport (GP): The dataset was collected over the University of Southern Mississippi’s Gulfpark Campus \citep{gader2013muufl}. The image consists of 72 bands, covering wavelengths from 0.37 to 1.04 $\upmu$m, with a spatial resolution of 1.0 meter. The dimensions of the Gulfpark (GP) image are $185 \times 89$ pixels. Figure~\ref{fig:GP_Results}a shows an RGB color composite of the scene, while the reference classification map in Figure~\ref{fig:GP_Results}b displays six classes. 

    \item Xuzhou (XZ): This dataset was collected by an airborne HYSPEX hyperspectral camera over the Xuzhou peri-urban site in November 2014 \citep{wang2019caps}. The dataset includes 436 spectral bands, covering wavelengths from 4.15 to 2.508 $\upmu$m, with noisy bands removed prior to experimentation. It consists of $500 \times 260$ pixels with a high spatial resolution of 0.73 meters per pixel. Figure \ref{fig:XZ_Results}a presents the pseudocolor composite image along with the labeled categories in Figure \ref{fig:XZ_Results}b. As with the previous datasets, unassigned pixels in the image are shown in black.
\end{enumerate}

\subsection{Experimental Setup and Evaluation Metrics} \label{sec:SetupAndMetrics}
Throughout the experiment, training samples were selected randomly, with only 1\% of the data utilized for model training (except for the SA dataset, where 0.5\% was used). The Adam optimizer was employed with a learning rate of $1 \times 10^{-3}$, dynamically adjusting the update step based on the mean and non-central variance of the gradients during backpropagation. Training was conducted in batches of 32, with a maximum limit of 150 epochs. To prevent overfitting, an early stopping strategy was implemented, ceasing training if no improvement was observed over 10 consecutive epochs, at which point the model reverted to its best-performing weights.

The experiments were conducted on a Windows 10 PC equipped with an i7-9700K CPU, NVIDIA GeForce RTX 2080 GPU, and 32 GB RAM, using Python 3.9 and TensorFlow 2.10.0. To assess the classification performance of the proposed network, five quantitative metrics were employed: {overall accuracy (OA), average accuracy (AA), the kappa coefficient, {T}-statistic, and P-value. OA measures the proportion of correctly classified pixels across the entire dataset, while AA computes the mean accuracy across all individual classes. The kappa coefficient evaluates the agreement between predicted and true labels, accounting for chance agreement, with values ranging from 0 to 1. The {T}-statistic quantifies the standardized difference in mean accuracy between the proposed model and other methods relative to data variability—higher absolute values indicate stronger distinction. The P-value, derived from the {T}-statistic, estimates the probability that the observed performance difference occurred by chance. A low p-value (typically < 0.05) supports the statistical significance of the model’s superiority, while a high p-value implies that the difference may not be meaningful.}

\subsection{Parameter Analysis}
Hyperparameters are pivotal in determining model performance, and selecting optimal values is essential for effective training. This study explores how patch size and the number of principal components impact model accuracy across various HSI datasets. Here, patch size refers to the spatial area within an image patch that contributes information for HSI classification. Larger patch sizes may inadvertently capture data from neighboring classes, which could hinder feature extraction, while smaller patches might limit the spatial context, reducing classification accuracy. To examine these effects, patch sizes of \{$5 \times 5$, $7 \times 7$, $9 \times 9$, $11 \times 11$, $13 \times 13$, $15 \times 15$, and $17 \times 17$\} were tested. In terms of spectral dimensions, the number of principal components was varied \{10, 15, 20, 25, 30, and 35\}, with initial increases generally improving performance, although accuracy eventually declined due to redundancy beyond an optimal level.

\begin{figure*}[!bt]
\centering
\includegraphics[width=0.75\linewidth]{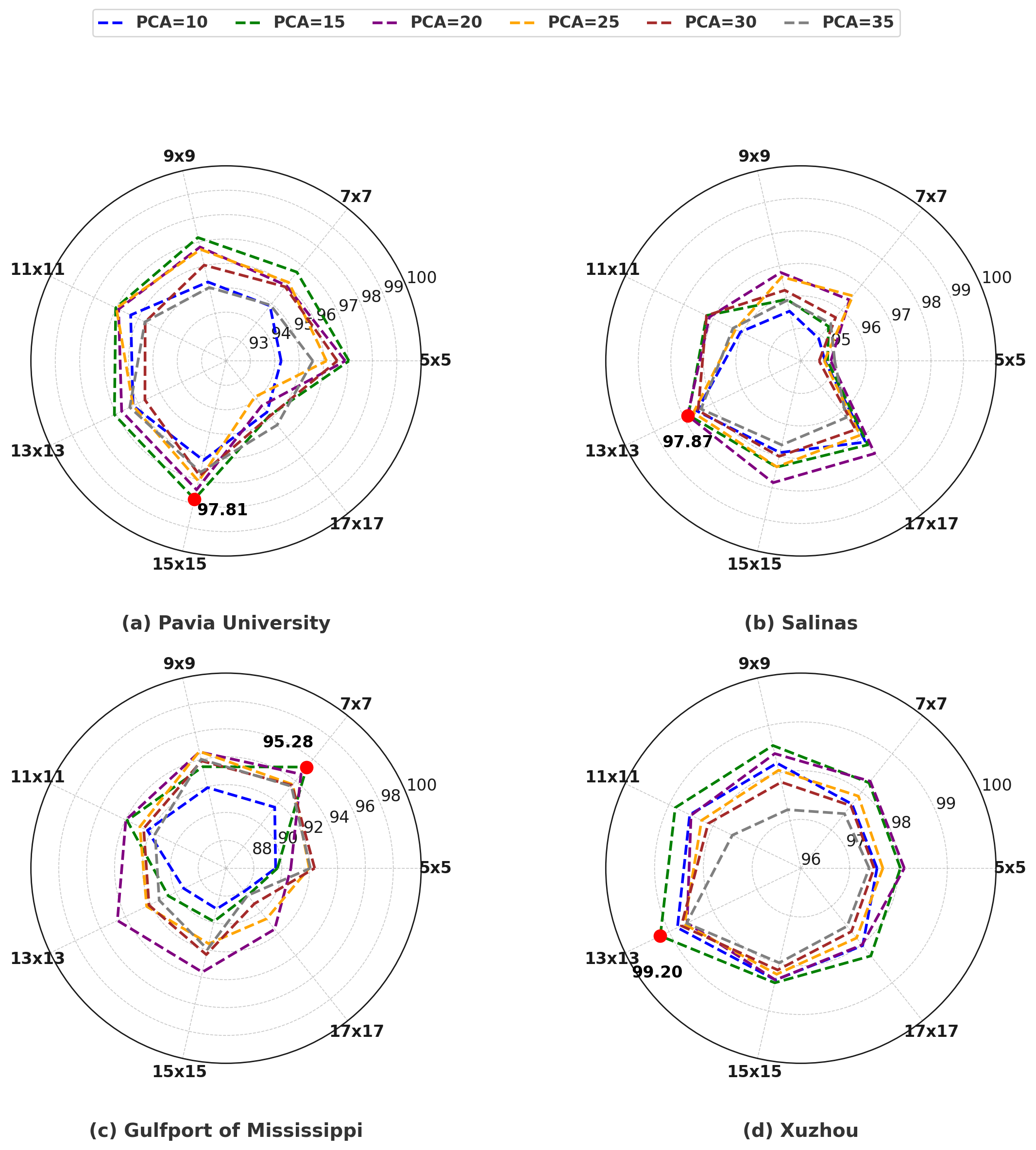}
\caption{Parameter Radar (Spider) plots of Overall Accuracy on four datasets. (a) Pavia University. (b) Salinas. (c) Gulfport of Mississippi. (d) Xuzhou.}
 \label{fig:PCA_vs_win}
\end{figure*}

Figure \ref{fig:PCA_vs_win} provides radar plots that illustrate classification accuracy across combinations of patch sizes and principal component (PCA) counts for the four datasets. For each dataset, specific configurations of patch size and PCA count yielded optimal results. The highest accuracies were achieved with the following configurations: $15 \times 15$ patch size with PCA = 15 for Pavia University (97.81\%), $13 \times 13$ with PCA = 20 for Salinas (99.22\%), $7 \times 7$ with PCA = 15 for Gulfport (95.28\%), and $13 \times 13$ with PCA = 15 for Xuzhou (99.20\%).

The Gulfport dataset, in particular, showed substantial fluctuations in accuracy with varying patch sizes and PCA counts, likely due to the model being trained on a limited dataset (1\% of the available data, equivalent to 60 training samples), highlighting the sensitivity to parameter adjustments under constrained data conditions. Overall, these findings underscore the importance of fine-tuning both patch size and PCA count based on dataset characteristics to achieve optimal classification performance. Table \ref{tab:winVsPCA} shows a summary of the Optimal Patch size along with the number of PCA count.

\begin{table}[tb!]
\centering
\caption{Optimal Patch Sizes and PCA Counts}
\label{tab:winVsPCA}
\begin{tabular}{ccc}
\hline
\textbf{Dataset} & \textbf{Optimal Patch Size} & \textbf{Optimal PCA Count}   \\
\hline
PU & $15 \times 15$ & 15  \\
SA & $13 \times 13$ & 20 \\
GP & $7 \times 7$ & 15  \\
XZ & $13 \times 13$ & 15  \\
\hline
\end{tabular}
\end{table}

\subsection{Impact of Attention mechanism}
In this section, we study the impact of each attention block discussed in Section \ref{sec:attention}. Five architectures were analyzed and explored in this research: the Baseline Mixer Module (MixerNet) without attention, MixerNet with Coordinate Attention (MixerCA), MixerNet with Convolutional Block Attention Module (MixerCBAM), MixerNet with Efficient Channel Attention (MixerECA), and MixerNet with Squeeze-and-Excitation (MixerSE). These models were rigorously trained and tested using the Pavia University dataset discussed in Section \ref{sec:dataset}. The evaluation metrics specified in Section \ref{sec:SetupAndMetrics} were employed to assess and compare the effectiveness of each trained model. Table \ref{tab:attentions} provides a summary of the performance metrics for the baseline model with different attention mechanisms.


The results in Table \ref{tab:attentions} demonstrate that attention mechanisms enhance MixerNet’s performance across OA, AA, and Kappa. MixerCA achieved the highest OA (98.23\%) and AA (96.44\%), indicating that Coordinate Attention significantly enhances classification accuracy among the variants. MixerSE also exhibited strong performance, with an OA of 98.01\% and an AA of 95.93\%, showing considerable improvement over the baseline MixerNet, which recorded an OA of 97.67\% and the lowest AA at 94.89\%. The Kappa scores, reflecting classification consistency, were highest for MixerCA (97.65) and MixerSE (97.36), underscoring the positive impact of attention mechanisms, particularly Coordinate Attention and Squeeze-and-Excitation, on optimizing MixerNet's performance. 

{To further ensure the generalizability of the Coordinate Attention module, we conducted a similar comparative study on the Gulfport of Mississippi dataset, as summarized in Table} \ref{tab:attentions2}. {MixerCA again achieved the highest OA (95.18\%) and AA (91.75\%), along with the top Kappa score (93.46), outperforming the MixerNet baseline and other attention-based variants. MixerSE followed closely with an OA of 95.00\%, AA of 91.60\%, and Kappa of 93.23. These consistent results across two distinct datasets reinforce the robustness and effectiveness of the CA module in enhancing classification performance.}

\begin{table}[tb!]
\centering
\caption{Performance metrics with different attention mechanisms on Pavia University Dataset.}
\label{tab:attentions}
\resizebox{8cm}{!} {
\begin{tabular}{lccc}
\hline
\multicolumn{1}{c}{} & \textbf{OA (\%)} & \textbf{AA (\%)} & \textbf{Kappa$\times$100} \\ \hline
MixerNet             & 97.67            & 94.89            & 96.91              \\
MixerCA              & 98.23            & 96.44            & 97.65              \\
MixerCBAM            & 97.69            & 95.43           & 96.93              \\
MixerECA             & 97.76            & 95.42            & 97.02              \\
MixerSE              & 98.01            & 95.93            & 97.36              \\ \hline
\end{tabular}
}
\end{table}

\begin{table}[tb!]
\centering
\caption{Performance metrics with different attention mechanisms on Gulfport of Mississippi Dataset.}
\label{tab:attentions2}
\resizebox{8cm}{!} {
\begin{tabular}{lccc}
\hline
\multicolumn{1}{c}{} & \textbf{OA (\%)} & \textbf{AA (\%)} & \textbf{Kappa$\times$100} \\ \hline
MixerNet             & 94.12            & 91.02            &  92.05             \\
MixerCA              & 95.18            & 91.75             & 93.46            \\
MixerCBAM            &  94.53           & 89.48           &  92.57             \\
MixerECA             &  94.35          & 90.11            & 92.36              \\
MixerSE              & 95.00            & 91.60            & 93.23              \\ \hline
\end{tabular}
}
\end{table}

\subsection{Comparison with Other Methods}


To assess the performance of the proposed model, this section presents a comparative analysis with several machine and deep learning methods developed in recent years, including SVM \citep{melgani2004classification}, MLP \citep{thakur2017hyper}, 2D-CNN \citep{chen2016deep}, 3D-CNN \citep{hamida20183}, Tri-CNN \citep{alkhatib2023tri}, PMI-CNN \citep{zhong2022hyperspectral}, and HybridSN \citep{roy2020}. Additionally, as vision transformers have demonstrated superior performance in classification tasks, we include comparisons with the Vision Transformer (ViT) \citep{dosovitskiy2020image} and Swin Transformer (Swin T.) \citep{liu2021swin, liu2023spectral} to highlight the effectiveness of our model against state-of-the-art approaches. {All models were implemented using their optimized hyperparameters as reported in their original publications to ensure a fair and representative evaluation.}

To ensure reliable classification results and minimize the impact of random sample selection, each experiment was repeated 10 times. The final results, expressed as the mean and variance, were obtained by averaging across these runs. Furthermore, detailed classification metrics were provided for each category. Quantitative comparisons of the methods are presented in Tables \ref{tab:PU_Results}--\ref{tab:XZ_Results}, with the best results in each table highlighted in bold. {This uniform configuration across models ensures that performance differences can be attributed to the model design rather than inconsistent training setups.}


\subsubsection{\highlighting{Model Size and Computational Efficiency}}
The \highlighting{model size and computational efficiency} of each model is summarized in Table \ref{tab:complexity}, listing the number of parameters, floating point operations (FLOPs), and multiply-and-accumulate operations (MACs) for comparison. \highlighting{The number of parameters reflects memory requirements, while FLOPs and MACs indicate computational cost.} Among the models, Tri-CNN has the highest \highlighting{values}, with 130,222,665 parameters, 260,116,992 FLOPs, and 130,058,496 MACs, reflecting significant computational demands. In contrast, the proposed MixerCA model, with 59,889 parameters, 19,145,472 FLOPs, and 9,318,144 MACs, achieves a favorable balance between performance and efficiency, outperforming simpler models such as MLP (137,993 parameters, 274,432 FLOPs) and 2D-CNN (83,401 parameters, 4,846,144 FLOPs).

Additionally, while PMI-CNN and HybridSN demonstrate high computational requirements with 104,416,512 and 97,483,008 FLOPs, respectively, MixerCA maintains lower FLOPs and MACs while integrating sophisticated components like depth-wise CNN, 2D-CNN and Coordinate Attention. ViT and Swin Transformer, with FLOPs of 13,765,120 and 3,568,712, respectively, are also \highlighting{relatively efficient but do not provide the same feature extraction capabilities as MixerCA}. This analysis highlights that MixerCA \highlighting{offers an efficient trade-off, combining reduced computational cost with strong classification performance} relative to both simpler architectures, such as MLP and 3D-CNN, and more demanding networks like Tri-CNN.

\begin{table}[tb!]
\centering
\caption{Parameters, FLOPs, and MACs of each Model used in the research}
\label{tab:complexity}
\resizebox{8cm}{!} {
\begin{tabular}{lccc}
\hline
\textbf{Model} & \textbf{Parameters} & \textbf{FLOPs} & \textbf{MACs} \\ \hline
MLP            & 137,993             & 274,432        & 137,216       \\
2D-CNN         & 83,401              & 4,846,144      & 2,419,200     \\
3D-CNN         & 73,417              & 35,493         & 16,960        \\
Tri-CNN        & 130,222,665         & 260,116,992    & 130,058,496   \\
PMI-CNN        & 3,779,593           & 104,416,512    & 52,208,256    \\
HybridnSN      & 2,313,465           & 97,483,008     & 48,741,504    \\
ViT            & 7,499,977           & 13,765,120     & 6,845,696     \\
Swin T.        & 140,425             & 3,568,712      & 1,659,832     \\
MixerCA        & 59,889              & 19,145,472     & 9,318,144     \\ \hline
\end{tabular}
}
\end{table}

\subsubsection{Results of Pavia University Dataset}

\begin{table*}[tb!]
\centering
\caption{Experimental Results of different methods on Pavia University Dataset.}
\label{tab:PU_Results}
\resizebox{\linewidth}{!} {
\begin{tabular}{ccc|cccccccccc}
\hline
\textbf{Class} & \textbf{Train} & \textbf{Test} & \textbf{SVM} & \textbf{MLP} & \textbf{2D-CNN} & \textbf{3D-CNN} & \textbf{Tri-CNN} & \textbf{PMI-CNN} & \textbf{HybridnSN} & \textbf{ViT} & \textbf{Swin T.} & \textbf{MixerCA}        \\ \hline
1              & 66             & 6,565          & 82.97        & 88.22        & 93.70            & 94.45           & 93.24            & 91.13            & \textbf{98.25}     & 89.37        & 93.26            & 96.06                   \\
2              & 186            & 18,463        & 93.06        & 97.25        & 99.46           & 99.10            & \textbf{100.00}     & 99.10             & 99.91              & 99.79        & 98.69            & 99.84                   \\
3              & 21             & 2,078         & 60.41        & 59.17        & 74.37           & 88.42           & 81.85            & 62.08            & \textbf{90.00}        & 71.46        & 74.94            & 82.47                   \\
4              & 31             & 3,033         & 85.31        & 82.87        & 89.88           & 90.54           & 92.89            & 91.29            & 85.87              & 87.83        & 80.29            & \textbf{96.57}          \\
5              & 13             & 1,332         & 99.55        & 95.46        & 99.03           & \textbf{99.93}  & 99.93            & 89.89            & 98.88              & 98.14        & 96.73            & 98.96                   \\
6              & 50             & 4,979         & 77.99        & 66.06        & 72.10            & 90.40            & 93.80             & 96.12            & 92.34              & 84.41        & 87.97            & \textbf{99.98}          \\
7              & 13             & 1,317         & 75.34        & 47.44        & 94.14           & 99.47           & 92.41            & 97.97            & 98.27              & 89.40         & 98.05            & \textbf{99.77}          \\
8              & 37             & 3,645         & 78.63        & 65.64        & 73.19           & 83.98           & 92.10             & 83.27            & 84.71              & 80.39        & 80.66            & \textbf{97.42}          \\
9              & 10             & 937           & \textbf{100.00} & 93.56        & 84.90            & 86.59           & 98.31            & 91.66            & 89.55              & 85.64        & 75.71            & 93.66                   \\ \hline
\multicolumn{3}{c}{OA   (\%)}                   & 86.13$\pm$3.12        & 84.88$\pm$2.65         & 90.67$\pm$2.03           & 94.68$\pm$1.97            & 95.87$\pm$1.32             & 93.29$\pm$1.98            & 95.65$\pm$1.49              & 91.76$\pm$1.86         & 91.96$\pm$1.56           & \textbf{97.81$\pm$0.42} \\
\multicolumn{3}{c}{AA   (\%)}                   & 83.94$\pm$4.35        & 78.05$\pm$3.62         & 87.14$\pm$2.96            & 92.76$\pm$2.72            & 94.04$\pm$1.45             & 89.58$\pm$2.21             & 93.34$\pm$1.86               & 87.82$\pm$2.48         & 87.83$\pm$2.32             & \textbf{96.25$\pm$1.15}          \\
\multicolumn{3}{c}{Kappa   x100}                & 81.57$\pm$3.26        & 79.56$\pm$2.96         & 87.44$\pm$2.55            & 92.93$\pm$1.23            & 94.50$\pm$1.12              & 91.07$\pm$1.85             & 94.19$\pm$1.69              & 89.00$\pm$1.99            & 89.27$\pm$2.01             & \textbf{97.10$\pm$0.75}           \\ \hline

\multicolumn{3}{c}{{T}-Statistic} & 11.73 & 15.24 & 10.89 & 4.91 & 4.43 & 7.06 & 4.41 & 10.03 & 11.45 & - \\
\multicolumn{3}{c}{P-Value} & 6.81$\times$10$^{-7}$ & 5.71$\times$10$^{-8}$ & 8.82$\times$10$^{-7}$ & 6.45$\times$10$^{-4}$ & 1.06$\times$10$^{-3}$ & 3.81$\times$10$^{-5}$ & 1.19$\times$10$^{-3}$ & 1.65$\times$10$^{-6}$ & 3.47$\times$10$^{-7}$ & - \\
 \hline

\end{tabular}
}

\end{table*}
\begin{figure*}[!bt]
\centering
\includegraphics[width=0.85\linewidth]{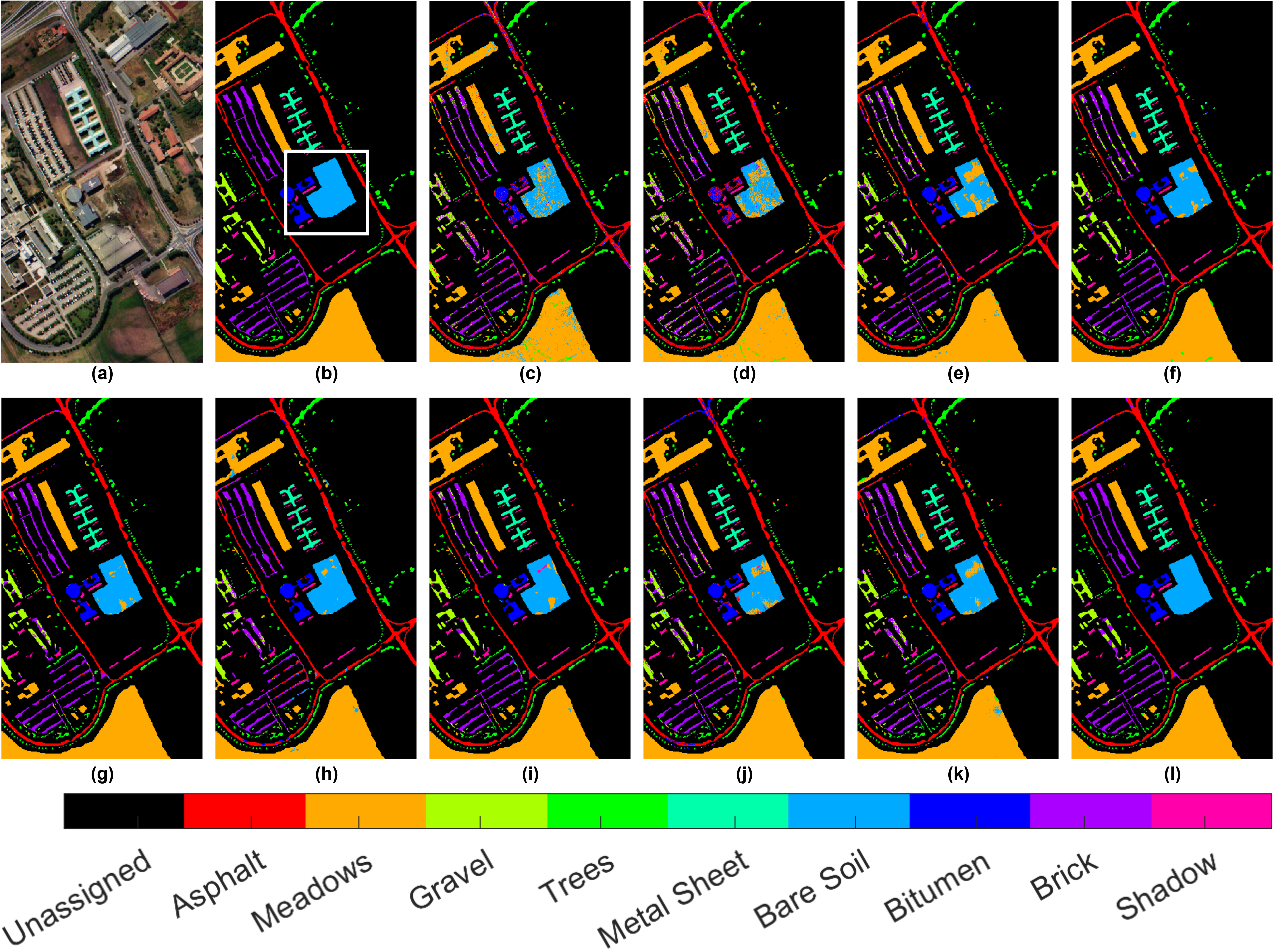}
\caption{{Classification results for the Pavia University dataset. (a) RGB image and (b) reference ground truth map are shown for visual context. Subfigures (c)–(l) present the classification outputs of various models, including traditional machine learning methods (SVM, MLP), CNN-based models (2D-CNN, 3D-CNN, Tri-CNN, PMI-CNN, HybridSN), transformer-based models (ViT, Swin Transformer), and the proposed MixerCA. The figure highlights the visual differences in classification quality and spatial consistency across methods, illustrating the superior performance of MixerCA in preserving structural details and reducing misclassifications, especially in the \textit{Bare Soil} Class}.}
 \label{fig:PU_Results}
\end{figure*}

Table \ref{tab:PU_Results} summarizes the classification accuracy of each method on the PU dataset, with the proposed MixerCA model demonstrating the best performance, achieving an overall accuracy (OA) of 97.81\%. In contrast, the SVM model showed a comparatively low OA of 86.13\%, primarily due to its reliance solely on spectral information, which limits its ability to capture spatial patterns critical for accurate classification. The MLP model, also lacking explicit spatial feature extraction, achieved an even lower OA of 84.88\%. The 2D-CNN model, using 2D filters for spatial feature extraction, improved the OA to 90.67\%, while the 3D-CNN model reached 94.68\% by jointly capturing spatial and spectral features with 3D filters. The Tri-CNN further optimized spatial-spectral feature extraction, resulting in an OA of 95.87\%, slightly outperforming PMI-CNN, which achieved an OA of 93.29\% despite using four parallel 2D convolutional branches for feature fusion. The HybridSN model, which integrates three layers of 3D-CNN with one layer of 2D-CNN to capture both spectral-spatial and spatial features, achieved an OA of 95.65\%. Transformer-based models, such as ViT and Swin Transformer, attained OAs of 91.76\% and 91.96\%, respectively, benefiting from attention mechanisms to capture global dependencies. However, these transformer-based models have a major drawback: they require large training datasets to perform optimally, which can limit their effectiveness on smaller datasets. The proposed MixerCA model not only achieved the highest OA but also excelled in class-specific accuracy across four key classes (Classes 4, 6, 7, and 8), exhibiting strong spatial coherence and close alignment with the ground truth.

\begin{table*}[tb!]
\caption{Experimental Results of different methods on Salinas Dataset.}
\centering

\label{tab:SA_Results}
\resizebox{\linewidth}{!} {
\begin{tabular}{ccc|cccccccccc}
\hline
\textbf{Class} & \textbf{Train} & \textbf{Test} & \textbf{SVM}   & \textbf{MLP}   & \textbf{2D-CNN} & \textbf{3D-CNN} & \textbf{Tri-CNN} & \textbf{PMI-CNN} & \textbf{HybridnSN} & \textbf{ViT}    & \textbf{Swin T.} & \textbf{MixerCA}        \\ \hline
1              & 10             & 1999          & 99.35          & 99.45          & \textbf{100.00} & \textbf{100.00} & \textbf{100.00}  & 99.65            & \textbf{100.00}    & 99.95           & \textbf{100.00}  & \textbf{100.00}         \\
2              & 19             & 3707          & 99.54          & 99.60          & \textbf{100.00} & \textbf{100.00} & \textbf{100.00}  & \textbf{100.00}  & 99.81              & 99.97           & 99.92            & \textbf{100.00}         \\
3              & 10             & 1966          & 97.72          & 90.33          & 99.49           & 95.60           & 93.67            & 98.89            & 99.60              & \textbf{100.00} & 99.85            & 99.90                   \\
4              & 7              & 1387          & 97.99          & 98.42          & 85.15           & 96.63           & 82.78            & 95.12            & 92.90              & 77.12           & 96.41            & \textbf{98.92}          \\
5              & 13             & 2665          & 95.67          & 88.83          & 97.35           & 97.31           & 95.07            & 98.06            & \textbf{100.00}    & 99.96           & 98.62            & 99.93                   \\
6              & 20             & 3939          & 99.49          & 99.82          & \textbf{100.00} & \textbf{100.00} & 99.97            & \textbf{100.00}  & \textbf{100.00}    & 99.90           & 99.87            & \textbf{100.00}         \\
7              & 18             & 3561          & 99.66          & 99.47          & 99.86           & 95.89           & 99.86            & \textbf{99.97}   & 99.94              & 97.88           & 99.89            & 99.83                   \\
8              & 56             & 11215         & 71.55          & 73.26          & 84.43           & 88.31           & 86.88            & 82.76            & 87.84              & 84.86           & 84.94            & \textbf{95.87}          \\
9              & 31             & 6172          & 99.24          & 99.82          & \textbf{100.00} & \textbf{100.00} & 99.61            & \textbf{100.00}  & \textbf{100.00}    & \textbf{100.00} & \textbf{100.00}  & \textbf{100.00}         \\
10             & 16             & 3262          & 89.29          & 84.90          & 97.28           & 93.44           & 95.52            & 97.77            & 97.80              & 98.11           & \textbf{99.91}   & 98.20                   \\
11             & 5              & 1063          & 76.31          & 86.89          & 92.32           & 93.54           & 93.91            & 81.27            & \textbf{99.63}     & 72.57           & 98.97            & 98.88                   \\
12             & 10             & 1917          & 99.27          & 93.31          & 99.41           & 97.66           & 79.45            & 97.15            & 97.87              & 99.07           & 97.87            & \textbf{99.43}          \\
13             & 5              & 911           & 88.65          & 83.52          & 98.36           & 97.60           & \textbf{100.00}  & 69.65            & 20.74              & 50.11           & 93.23            & 91.16                   \\
14             & 5              & 1065          & 76.17          & 80.75          & 94.30           & 97.85           & 22.71            & 99.16            & 94.11              & \textbf{98.79}  & 88.79            & 95.70                   \\
15             & 36             & 7232          & 51.82          & 63.46          & 72.56           & 83.64           & 81.04            & 79.88            & 90.18              & 91.98           & 82.00            & \textbf{93.13}          \\
16             & 9              & 1798          & 96.79          & 93.08          & 97.40           & 99.45           & 99.89            & 98.78            & 98.06              & 94.19           & 92.97            & \textbf{99.94}          \\ \hline
\multicolumn{3}{c|}{OA (\%)}                    & 85.57$\pm$2.23 & 86.43$\pm$1.41 & 91.64$\pm$1.34  & 94.01$\pm$1.25  & 91.09$\pm$1.11   & 92.26$\pm$0.62   & 94.20$\pm$0.92     & 93.28$\pm$1.13  & 93.59$\pm$1.05   & \textbf{97.87$\pm$0.36} \\
\multicolumn{3}{c|}{AA (\%)}                    & 89.89$\pm$2.32 & 89.37$\pm$2.01 & 94.81$\pm$1.53  & 96.06$\pm$1.27  & 89.40$\pm$1.32   & 93.63$\pm$1.21   & 92.41$\pm$1.32     & 91.53$\pm$1.23  & 95.83$\pm$1.93   & \textbf{98.18$\pm$0.41} \\
\multicolumn{3}{c|}{Kappa x100}                 & 83.27$\pm$1.32 & 84.34$\pm$1.24 & 91.80$\pm$1.37  & 93.34$\pm$1.13  & 90.08$\pm$1.04   & 91.38$\pm$1.17   & 93.54$\pm$0.84     & 92.52$\pm$1.04  & 92.86$\pm$0.97   & \textbf{97.53$\pm$0.48} \\ \hline

\multicolumn{3}{c}{{T}-Statistic} & 17.2192 & 24.8596 & 14.1988 & 9.3837 & 18.3734 & 24.7446 & 11.7474 & 12.2389 & 12.1933 & - \\
\multicolumn{3}{c}{P-Value}       & 2.34$\times$10$^{-7}$ & 1.16$\times$10$^{-9}$ & 3.08$\times$10$^{-7}$ & 6.75$\times$10$^{-6}$ & 1.18$\times$10$^{-7}$ & 1.24$\times$10$^{-9}$ & 1.52$\times$10$^{-6}$ & 1.21$\times$10$^{-6}$ & 1.25$\times$10$^{-6}$ & - \\ \hline

\end{tabular}
}

\end{table*}
\begin{figure*}[!bt]
\centering
\includegraphics[width=0.85\linewidth]{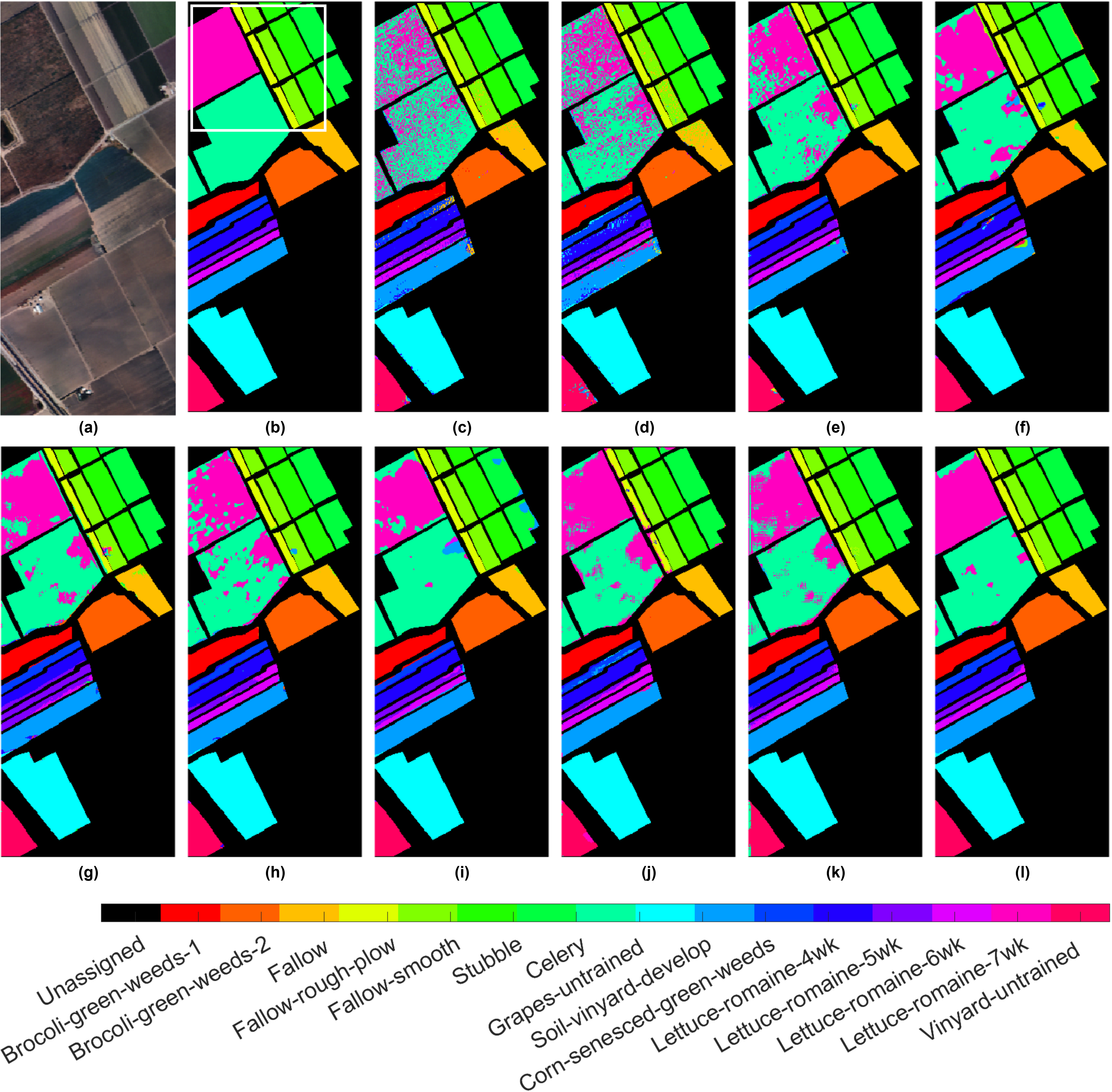} 
\caption{{Classification results for the Salinas dataset. (a) RGB image and (b) reference ground truth map are shown for visual reference. Subfigures (c)–(l) display the classification maps generated by different models, including classical machine learning approaches (SVM, MLP), deep CNN-based models (2D-CNN, 3D-CNN, Tri-CNN, PMI-CNN, HybridSN), transformer-based methods (ViT, Swin Transformer), and the proposed MixerCA. The visual comparison demonstrates the superior spatial coherence and classification accuracy achieved by MixerCA, particularly in the upper left region of the scene, where class boundaries are better preserved and misclassifications are noticeably reduced}.}
 \label{fig:SA_Results}
\end{figure*}

For a visual comparison of the results, Figure \ref{fig:PU_Results} displays the classification maps produced by each method alongside the reference map. As shown in Figure \ref{fig:PU_Results}c, the SVM method produces a large number of misclassified pixels, resulting in low classification accuracy. In contrast, the classification map generated by the proposed MixerCA model most closely resembles the reference map, especially in the region within the white box (representing the Bare Soil class), where MixerCA demonstrates high accuracy and spatial coherence.

{To further validate the effectiveness of the proposed MixerCA model, a paired {t}-test was conducted to statistically compare its overall accuracy against each competing method on the Pavia University dataset. The final two rows in Table}~\ref{tab:PU_Results} {report the corresponding {T}-Statistic and {P}-Value for each comparison. The results indicate that MixerCA significantly outperforms all baseline methods, with all {P}-Values falling below the 0.05 threshold, indicating strong statistical significance. In particular, comparisons with SVM, MLP, and 2D-CNN yielded very large {T}-Statistic values (11.73, 15.24, and 10.89, respectively), confirming substantial performance gaps. Even when compared with stronger models such as 3D-CNN, Tri-CNN, and HybridSN, the improvements remain statistically significant (e.g., {P} = 6.45$\times$10$^{-4}$ for 3D-CNN and {P} = 1.19$\times$10$^{-3}$ for HybridSN). These findings provide robust evidence that the superior classification performance of MixerCA is not due to chance but reflects genuine improvements in model capability.}

\subsubsection{Results of Salinas Dataset}

Table~\ref{tab:SA_Results} shows that the proposed MixerCA model achieved the highest Kappa score of 97.53, with a standard deviation of only 0.48, indicating both high accuracy and stability. In contrast, traditional methods such as SVM and MLP displayed not only lower Kappa values (83.27 and 84.34, respectively) but also higher variability, reflecting less consistent performance.

Among the CNN and Transformer-based models, CNN approaches such as 2D-CNN, 3D-CNN, and HybridSN show relatively stable Kappa scores (91.80 to 93.54) with standard deviations around 1.0–1.3, whereas Transformer models like ViT and Swin Transformer achieved Kappa values up to 94.69 with slightly lower variability. Nevertheless, MixerCA’s superior Kappa score and minimal standard deviation highlight its advantage in both performance and reliability on the Salinas dataset.

{The table also shows the T-statistic and P-values, demonstrating that MixerCA consistently outperforms all baseline methods. All comparisons yielded T-statistics greater than zero, and the corresponding P-values remained well below the 0.05 threshold, confirming the statistical significance of the improvements at a 95\% confidence level. Notably, large T-statistic values were observed for MLP (24.8596), PMI-CNN (24.7446), and Tri-CNN (18.3734), emphasizing the substantial performance gains of MixerCA over these methods.}

MixerCA achieved the highest accuracy in 9 out of 16 classes, with perfect scores in 4 of those classes, underscoring its effectiveness in capturing both spatial and spectral features across diverse class types. The visual comparison in Figure~\ref{fig:SA_Results} further supports these quantitative findings; within the region highlighted in the white box, MixerCA accurately delineates class boundaries, closely aligning with the ground truth.

\subsubsection{Results of Gulfport of Mississippi Dataset}

Table \ref{tab:GP_Results} presents the classification outcomes for various methods applied to the Gulfport of Mississippi dataset, where MixerCA stands out with the highest average accuracy (AA) of 91.60\%. This AA represents a notable improvement over the next best-performing model, HybridnSN, which achieved an AA of 81.01\%, indicating a substantial 10.59\% difference. This considerable margin reflects MixerCA's proficiency in accurately classifying across multiple classes while effectively capturing both spatial and spectral variations.

In contrast, traditional methods like SVM and MLP reached AAs of 74.34\% and 78.12\%, respectively, highlighting their limitations in feature extraction for this dataset. Among convolutional models, performance varied: 2D-CNN achieved an AA of 69.98\%, while 3D-CNN and HybridnSN performed better, with AAs of 81.84\% and 81.01\%, respectively. Transformer-based models, including ViT and Swin Transformer, achieved AAs of 76.34\% and 80.66\%, effectively capturing dependencies within the data but still falling short of MixerCA’s high performance.

{The table also includes T-statistic and P-value metrics to validate the statistical significance of MixerCA’s superiority. Positive T-statistic values across all comparisons confirm that MixerCA consistently outperforms competing methods. Moreover, all P-values are well below the 0.05 threshold, establishing statistical significance at the 95\% confidence level. Notable examples include SVM (T = 27.85, P = 1.62$\times$10$^{-12}$), MLP (T = 19.04, P = 3.00$\times$10$^{-10}$), and Tri-CNN (T = 14.42, P = 3.47$\times$10$^{-9}$), which emphasize the substantial and reliable performance gap in favor of MixerCA.}

\begin{figure*}[!tb]
\centering
\includegraphics[width=0.85\linewidth]{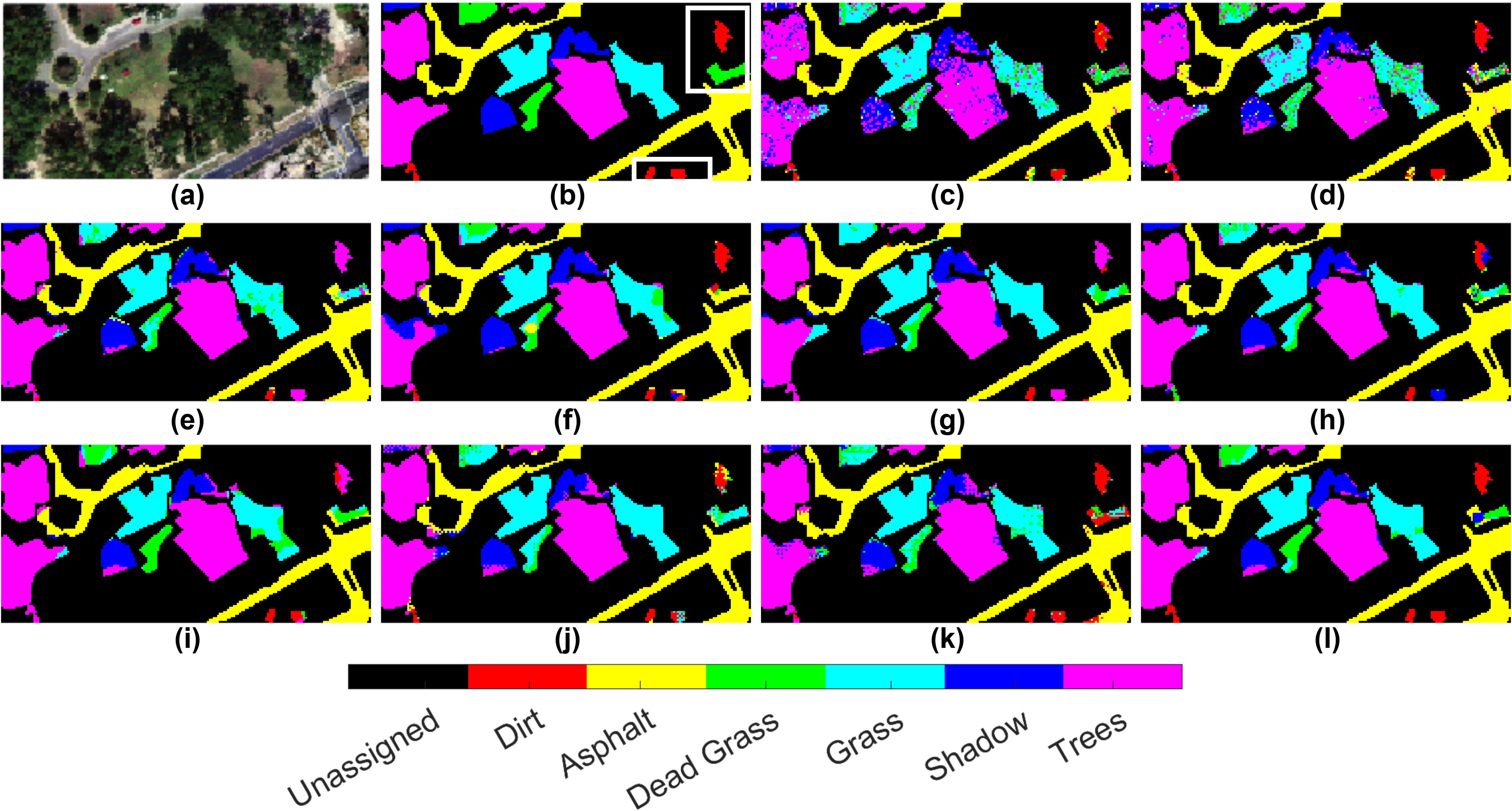}
\caption{{Classification results for the Gulfport of Mississippi dataset. (a) RGB image and (b) reference ground truth map are provided for visual context. Subfigures (c)–(l) show the classification outputs produced by various models, including traditional classifiers (SVM, MLP), CNN-based approaches (2D-CNN, 3D-CNN, Tri-CNN, PMI-CNN, HybridSN), transformer-based models (ViT, Swin Transformer), and the proposed MixerCA. Notably, the dataset includes a highly underrepresented \textit{Dirt} class with only two training samples, posing a significant challenge. Despite this, MixerCA demonstrates strong generalization capability, accurately identifying regions belonging to rare classes while maintaining spatial consistency across the scene.}}
 \label{fig:GP_Results}
\end{figure*}

Figure \ref{fig:GP_Results} illustrates the classification maps for each method, with MixerCA producing a map that most closely matches the reference data, particularly in the area within the white box (Class \#1). Remarkably, despite having only 2 training samples for this class, MixerCA achieved a perfect classification score of 100.00\%, well above the second-best model, Swin T., which scored only 82.29\%. Similarly, in Class \#2 (marked in yellow) with only 16 training samples, MixerCA achieved 99.88\%, demonstrating its strong performance even with limited training data. For Class \#3, which had only 4 training samples, MixerCA achieved a score of 68.04\%, placing it second to HybridnSN’s 75.34\%. In fact, MixerCA consistently ranked first or second across all classes, emphasizing its competitive and reliable performance across diverse classification challenges.

\begin{table*}[tb!]
\centering
\caption{Experimental Results of different methods on Gulfport of Mississippi Dataset.}.
\label{tab:GP_Results}
\resizebox{\linewidth}{!} {
\begin{tabular}{ccc|cccccccccc}
\hline
\textbf{Class} & \textbf{Train} & \textbf{Test} & \textbf{SVM} & \textbf{MLP} & \textbf{2D-CNN} & \textbf{3D-CNN} & \textbf{Tri-CNN} & \textbf{PMI-CNN} & \textbf{HybridnSN} & \textbf{ViT} & \textbf{Swin T.} & \textbf{MixerCA} \\ \hline
1              & 2              & 173           & 69.71        & 78.86        & 14.86           & 74.29           & 17.71            & 44.57            & 42.86              & 62.86        & 82.29            & \textbf{100.00}  \\
2              & 16             & 1,641         & 99.03        & 97.83        & 98.67           & 98.49           & 98.31            & 99.76            & 98.91              & 96.56        & 97.28            & \textbf{99.88}   \\
3              & 4              & 434           & 38.13        & 48.40        & 27.85           & 39.73           & 39.50            & 35.39            & \textbf{75.34}     & 26.48        & 36.07            & 68.04            \\
4              & 10             & 960           & 81.44        & 71.55        & 92.89           & 91.55           & \textbf{99.90}   & 98.35            & 87.01              & 98.87        & 94.95            & 97.42            \\
5              & 5              & 522           & 72.49        & 77.42        & 87.67           & 94.88           & \textbf{88.43}   & 88.24            & 82.92              & 75.52        & 77.23            & 86.34            \\
6              & 23             & 2,307         & 85.24        & 94.68        & 97.94           & 92.10           & 93.48            & 97.94            & \textbf{99.01}     & 97.77        & 96.14            & 97.90            \\ \hline
\multicolumn{3}{c}{OA (\%)}                     & 83.45$\pm$1.23 & 86.58$\pm$1.34 & 89.03$\pm$1.51 & 89.72$\pm$1.26 & 89.32$\pm$1.19 & 91.64$\pm$1.21 & 92.37$\pm$0.98 & 89.57$\pm$1.14 & 89.91$\pm$1.07 & \textbf{95.28$\pm$0.54}   \\
\multicolumn{3}{c}{AA (\%)}                     & 74.34$\pm$1.05 & 78.12$\pm$1.16 & 69.98$\pm$1.20 & 81.84$\pm$1.14 & 72.89$\pm$1.09 & 77.38$\pm$1.02 & 81.01$\pm$0.97 & 76.34$\pm$1.03 & 80.66$\pm$1.00 & \textbf{91.60$\pm$0.43}   \\
\multicolumn{3}{c}{Kappa x100}                  & 77.68$\pm$1.13 & 81.67$\pm$1.20 & 84.91$\pm$1.36 & 86.14$\pm$1.18 & 85.46$\pm$1.10 & 88.61$\pm$1.11 & 89.57$\pm$0.94 & 85.73$\pm$1.02 & 86.34$\pm$0.99 & \textbf{93.60$\pm$0.48}   \\ \hline

\multicolumn{3}{c}{{T}-Statistic} & 27.85 & 19.04 & 12.32 & 12.83 & 14.42 & 8.69 & 8.22 & 14.31 & 14.17 & - \\ 
\multicolumn{3}{c}{P-Value} & 1.62$\times$10$^{-12}$ & 3.00$\times$10$^{-10}$ & 6.93$\times$10$^{-8}$ & 1.92$\times$10$^{-8}$ & 3.47$\times$10$^{-9}$ & 1.23$\times$10$^{-6}$ & 9.89$\times$10$^{-7}$ & 2.86$\times$10$^{-9}$ & 2.08$\times$10$^{-9}$ & - \\ \hline

\end{tabular}
}

\end{table*}

\subsubsection{Results of Xuzhou Dataset}

Table \ref{tab:XZ_Results} presents the classification performance of various methods on the Xuzhou dataset, with MixerCA achieving the highest scores across all key metrics. MixerCA obtained an OA of 99.20\%, AA of 98.77\%, and a Kappa coefficient of 99.14, establishing it as the top-performing model for this dataset. Notably, MixerCA also achieved the highest classification accuracy in 5 out of the 9 classes, demonstrating its strong capability in handling both spatial and spectral complexities. In comparison, the second-best model, HybridnSN, achieved an OA of 97.38\%, an AA of 97.49\%, and a Kappa of 97.08, resulting in a gap of approximately 1.82\%, 1.28\%, and 2.06 in OA, AA, and Kappa, respectively. This margin underscores the superior performance of MixerCA in accurately capturing class-specific details and achieving reliable classification outcomes.

\begin{figure*}[!bt]
\centering
\includegraphics[width=0.85\linewidth]{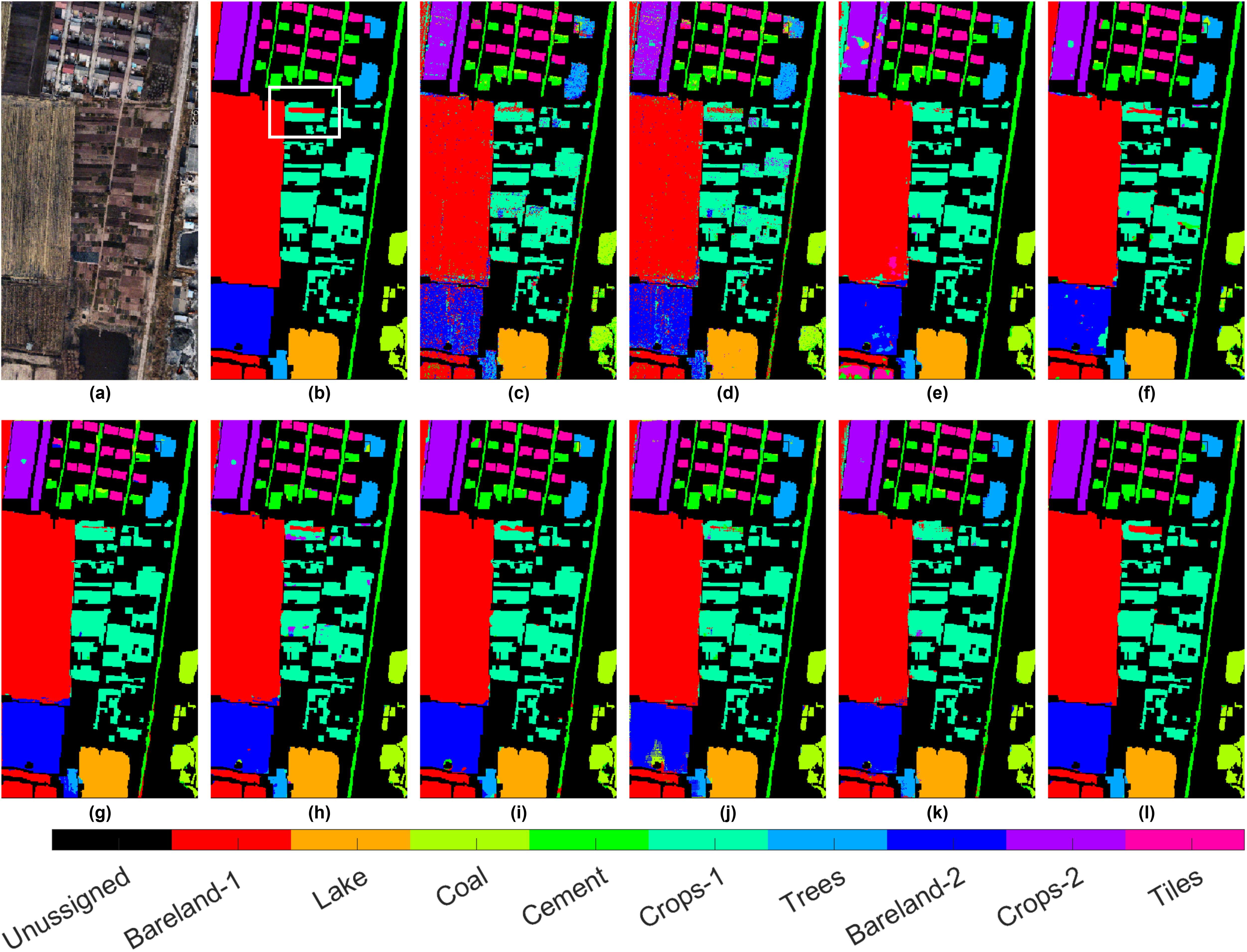}
\caption{{Classification results for the Xuzhou dataset. (a) RGB image and (b) reference ground truth map are included for visual reference. Subfigures (c)–(l) illustrate the classification outputs of various models, including conventional classifiers (SVM, MLP), CNN-based methods (2D-CNN, 3D-CNN, Tri-CNN, PMI-CNN, HybridSN), transformer-based models (ViT, Swin Transformer), and the proposed MixerCA. The visual results indicate that most of the classes were correctly identified using MixerCA, with improved spatial coherence and reduced misclassification compared to other methods}.}
 \label{fig:XZ_Results}
\end{figure*}

Other models exhibited a range of performance levels on the Xuzhou dataset. SVM and MLP achieved overall accuracies (OAs) of 88.57\% and 90.43\%, respectively, reflecting some limitations in managing the dataset's complex features. Convolutional models showed improved results, with 3D-CNN and PMI-CNN achieving OAs of 95.73\% and 95.53\%, respectively, while transformer-based models, ViT and Swin Transformer, attained OAs of 94.60\% and 95.23\%. In contrast, MixerCA achieved the highest metrics overall, demonstrating consistently strong performance across multiple classes. Figure \ref{fig:XZ_Results} further supports the results shown in Table \ref{tab:XZ_Results}, particularly in the area highlighted by the white box (Bareland-1).

\begin{table*}[tb!]
\caption{Experimental Results of different methods on Xuzhou Dataset.}
\centering

\label{tab:XZ_Results}
\resizebox{\linewidth}{!} {
\begin{tabular}{ccc|cccccccccc}
\hline
\textbf{Class} & \textbf{Train} & \textbf{Test} & \textbf{SVM}   & \textbf{MLP} & \textbf{2D-CNN} & \textbf{3D-CNN} & \textbf{Tri-CNN} & \textbf{PMI-CNN} & \textbf{HybridnSN} & \textbf{ViT} & \textbf{Swin T.} & \textbf{MixerCA} \\ \hline
1              & 263            & 26,133        & 93.93          & 94.72        & 89.42           & 98.14           & 97.00            & 97.73            & 98.01              & 97.53        & 96.72            & \textbf{99.16}   \\
2              & 40             & 3,987         & 97.77          & 97.29        & 97.99           & 99.03           & 98.78            & \textbf{99.60}   & 99.08              & 98.51        & 98.76            & 99.18            \\
3              & 28             & 2,755         & 77.87          & 86.92        & 91.88           & 88.14           & 97.31            & 98.02            & 96.51              & 98.92        & 95.65            & \textbf{98.96}   \\
4              & 52             & 5,162         & 86.98          & 85.86        & 95.28           & 95.86           & 91.41            & 97.62            & 97.53              & 91.41        & 94.99            & \textbf{98.14}   \\
5              & 132            & 13,052        & 93.45          & 93.70        & 99.36           & 95.81           & \textbf{99.68}   & 95.76            & 99.07              & 98.38        & 98.17            & 98.56            \\
6              & 24             & 2,412         & 64.78          & 70.94        & 92.73           & 97.37           & 85.34            & 97.37            & 95.32              & 90.76        & 96.39            & \textbf{99.67}   \\
7              & 70             & 6,920         & 77.35          & 84.59        & 96.22           & 94.55           & 98.78            & 98.96            & 98.21              & 90.77        & 98.51            & \textbf{99.64}   \\
8              & 48             & 4,729         & 95.81          & 90.94        & 89.28           & 97.91           & 97.80            & 99.00            & \textbf{100.00}    & 99.12        & 99.39            & 99.87            \\
9              & 31             & 3,039         & \textbf{97.33} & 93.62        & 95.67           & 93.97           & 94.46            & 96.19            & 93.78              & 96.64        & 95.24            & 95.34            \\ \hline
\multicolumn{3}{c}{OA (\%)}                     & 88.57$\pm$1.28 & 90.43$\pm$1.43 & 92.64$\pm$1.54 & 95.73$\pm$1.27 & 94.78$\pm$1.12 & 95.53$\pm$1.23 & 97.38$\pm$0.92 & 94.60$\pm$1.13 & 95.23$\pm$1.04 & \textbf{99.20$\pm$0.57} \\
\multicolumn{3}{c}{AA (\%)}                     & 87.38$\pm$1.07 & 88.90$\pm$1.14 & 94.05$\pm$1.26 & 95.65$\pm$1.13 & 95.53$\pm$1.02 & 97.58$\pm$0.97 & 97.49$\pm$1.03 & 95.66$\pm$1.09 & 97.09$\pm$0.98 & \textbf{98.77$\pm$0.46} \\
\multicolumn{3}{c}{Kappa x100}                  & 87.27$\pm$1.19 & 89.34$\pm$1.21 & 91.80$\pm$1.32 & 95.25$\pm$1.18 & 94.18$\pm$1.07 & 95.01$\pm$1.11 & 97.08$\pm$0.89 & 94.00$\pm$1.02 & 94.69$\pm$0.94 & \textbf{99.14$\pm$0.48} \\ \hline

\multicolumn{3}{c}{{T}-Statistic} & 23.99 & 18.02 & 12.63 & 7.88 & 11.12 & 8.56 & 5.32 & 11.49 & 10.59 & - \\
\multicolumn{3}{c}{P-Value} & 8.66$\times$10$^{-12}$ & 6.10$\times$10$^{-10}$ & 4.62$\times$10$^{-8}$ & 3.40$\times$10$^{-6}$ & 3.92$\times$10$^{-8}$ & 1.25$\times$10$^{-6}$ & 8.57$\times$10$^{-5}$ & 2.77$\times$10$^{-8}$ & 4.72$\times$10$^{-8}$ & - \\ \hline

\end{tabular}
}
\end{table*}

\subsubsection{Experimental Results of Other Datasets}
While the primary evaluation focuses on four benchmark datasets—Pavia University, Salinas, Gulfport of Mississippi, and Xuzhou—this section presents additional qualitative results on \highlighting{four} supplementary datasets to further explore the model's scalability and generalization. These results are not part of the main comparative analysis but offer supportive insights.

To this end, \highlighting{four} additional hyperspectral datasets acquired by different sensors were selected. The Augsburg (AU) dataset was collected using the HySpex airborne imaging spectrometer operated by the Remote Sensing Technology Institute of the German Aerospace Center \citep{hong2021multimodal}. The WHU-Hi-LongKou (LK) dataset was captured using the Headwall Nano-Hyperspec sensor in Longkou Town, Hubei Province, China \citep{zhong2020whu}. The Kansas (KN) dataset was obtained with the visible-SWIR AHSI sensor developed by the Shanghai Institute of Technical Physics, Chinese Academy of Sciences \citep{wang2022unified}. \highlighting{The Houston (HO) dataset was acquired by the ITRES CASI-1500 sensor over the University of Houston campus in June 2012, and was part of the 2013 IEEE GRSS Data Fusion Contest} \citep{debes2014hyperspectral}.

Table \ref{tab:others} presents the Kappa coefficient and overall classification accuracy for each algorithm on these \highlighting{four} datasets \highlighting{when trained with only 1\% of data}. The results show that the proposed model consistently achieves the highest accuracy across all datasets. However, the sub-optimal performances vary across algorithms, indicating that the proposed model provides robust feature extraction and demonstrates strong generalization across different types of data.

\begin{table}[tb!]
\centering
\caption{Overall Accuracy and Kappa coefficient for three datasets}
\label{tab:others}
\resizebox{0.6\linewidth}{!} {
\begin{tabular}{lcccccccc}
\hline
               & \multicolumn{2}{c}{\textbf{AU}}          & \multicolumn{2}{c}{\textbf{LK}}          & \multicolumn{2}{c}{\textbf{KN}}          & \multicolumn{2}{c}{\textbf{\highlighting{HO}}}          \\ \cline{2-9} 
\textbf{Model} & \textbf{OA (\%)} & \textbf{K$\times$100} & \textbf{OA (\%)} & \textbf{K$\times$100} & \textbf{OA (\%)} & \textbf{K$\times$100} & \textbf{OA (\%)} & \textbf{K$\times$100} \\ \hline
SVM            & 83.12            & 75.73                 & 96.45            & 95.33                 & 84.73            & 82.22                 & 82.55            & 81.11                 \\
MLP            & 84.77            & 77.94                 & 96.65            & 95.60                 & 83.04            & 80.44                 & 68.13            & 65.53                 \\
2D-CNN         & 91.38            & 87.58                 & 98.67            & 98.25                 & 85.51            & 82.29                 & 64.61            & 61.67                 \\
3D-CNN         & 91.88            & 88.36                 & 99.39            & 99.19                 & 88.50            & 85.96                 & 83.38            & 82.04                 \\
Tri-CNN        & 91.97            & 88.44                 & 99.44            & 99.27                 & 89.14            & 86.69                 & 73.23            & 71.15                 \\
PMI-CNN        & 92.90            & 89.76                 & 99.21            & 98.96                 & 88.91            & 85.67                 & 81.57            & 80.05                 \\
HybridSN       & 91.67            & 88.01                 & 99.22            & 98.98                 & 88.87            & 86.21                 & 79.88            & 78.25                 \\
ViT            & 91.93            & 88.39                 & 99.17            & 98.91                 & 85.69            & 82.67                 & 60.44            & 57.20                 \\
Swin T.        & 90.54            & 86.45                 & 98.79            & 98.42                 & 86.66            & 83.69                 & 70.99            & 68.58                 \\
MixerCA        & 93.62            & 90.87                 & 99.57            & 99.43                 & 91.83            & 90.35                 & 85.53            & 84.44                 \\ \hline
\end{tabular}
}
\end{table}

\subsection{{Model Performance Across Varying Training Data Percentages}}
\begin{figure*}[!bt]
\centering
\includegraphics[width=0.75\linewidth]{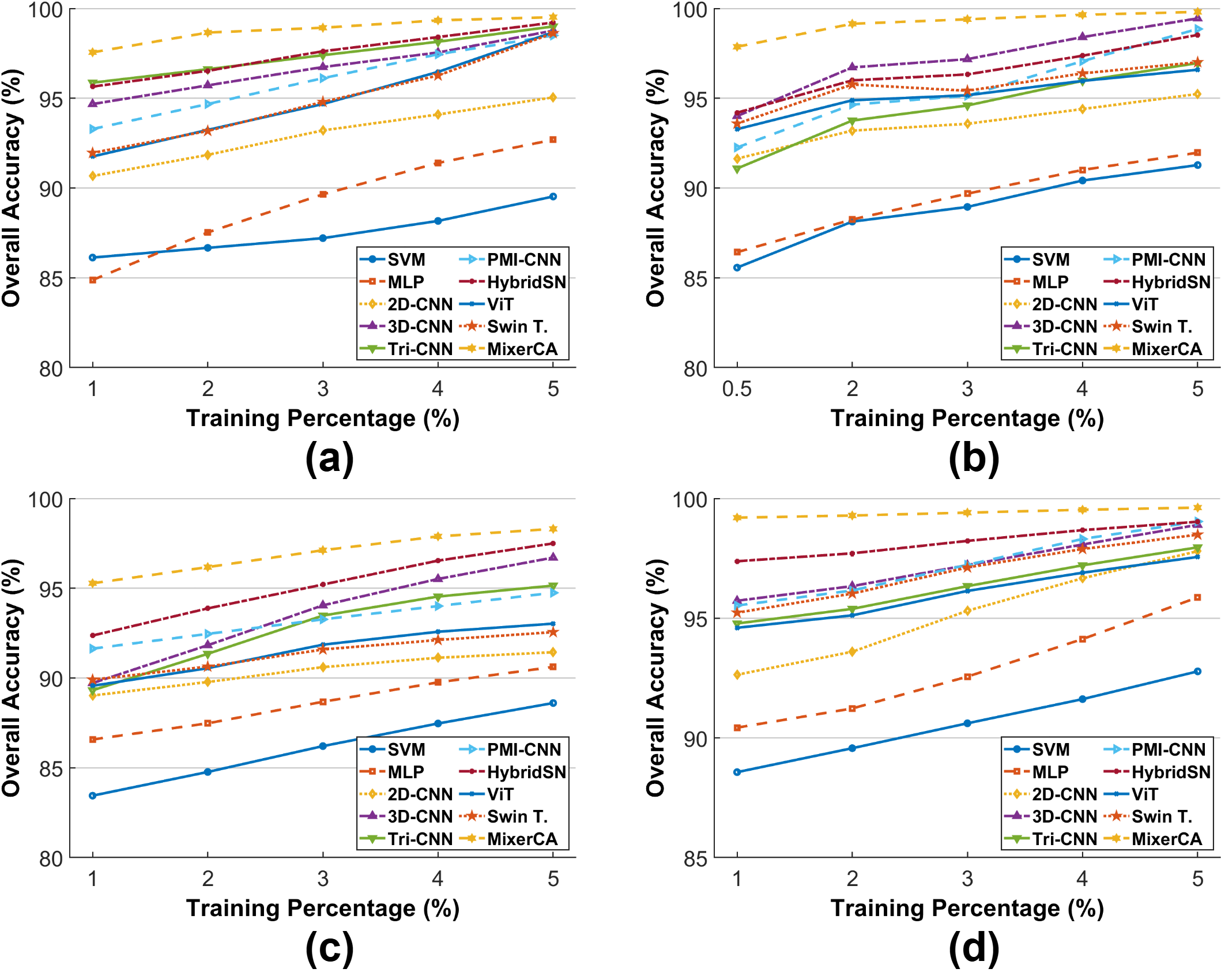}
\caption{Overall Accuracy with respect to training percentages: (a) Pavia University, (b) Salinas, (c) Gulfport of Mississippi, (d) Xuzhou.}
 \label{fig:precentages}
\end{figure*}

{To assess the generalization ability of the proposed MixerCA model under limited training data conditions, a comprehensive comparison was carried out against all baseline methods using varying training ratios from 1\% to 5\% per class (0.5\% to 5\% for the Salinas dataset). As shown in Figure}~\ref{fig:precentages}, {MixerCA consistently achieves superior performance across all scenarios. An overall increase in classification accuracy is observed with higher training percentages. While the Average Accuracy and Kappa metrics are not presented in the figure, they exhibit a similar upward trend consistent with the Overall Accuracy.}

{For the Pavia University dataset, MixerCA achieved the highest overall accuracy across all training ratios, reaching 97.56\% with only 1\% of training data and 99.52\% at 5\%. In comparison, HybridSN (95.65\% at 1\%, 99.22\% at 5\%) and ViT (91.76\% at 1\%, 98.68\% at 5\%) showed lower performance, particularly under low-data conditions. Although the performance gap narrows with more training data, MixerCA consistently maintained a slight advantage. These improvements were achieved with fewer parameters, highlighting the model’s efficiency as well as accuracy. Results are shown in Figure}~\ref{fig:precentages}(a).

{As shown in Figure}~\ref{fig:precentages}(b), {MixerCA also outperformed all competing methods on the Salinas dataset, achieving 97.87\% at 0.5\% training and 99.81\% at 5\%. Other models, such as HybridSN (94.20\% at 0.5\%, 98.53\% at 5\%) and Swin Transformer (93.59\% at 0.5\%, 97.01\% at 5\%), yielded lower accuracies, especially with limited samples. MixerCA maintained a consistent lead across all training ratios while requiring fewer parameters, reinforcing its robustness and efficiency.

For the Gulfport of Mississippi dataset, MixerCA achieved 95.28\% at 1\% training data and 98.31\% at 5\%, outperforming both CNN- and transformer-based models. Transformer models like ViT (89.57\% at 1\%, 93.03\% at 5\%) and Swin Transformer (89.91\% at 1\%, 92.56\% at 5\%) were less effective due to their reliance on large training sets. In contrast, MixerCA demonstrated strong performance even with limited annotations, surpassing HybridSN (92.37\% at 1\%, 97.50\% at 5\%) and 3D-CNN (89.72\% at 1\%, 96.71\% at 5\%), as illustrated in Figure~}\ref{fig:precentages}(c).

{On the Xuzhou dataset, MixerCA again achieved the best performance, with 99.20\% at 1\% training and 99.62\% at 5\%. While models like HybridSN (97.38\% at 1\%, 99.03\% at 5\%) and PMI-CNN (95.53\% at 1\%, 99.04\% at 5\%) performed well, MixerCA retained a clear edge. Transformer models such as ViT and Swin Transformer reached 97.56\% and 98.49\%, respectively, but did not surpass MixerCA. These results confirm the model’s generalization capability and efficiency under varying data availability, as shown in Figure~}\ref{fig:precentages}(d).

\section{Conclusion} \label{sec:conclusion}
In this paper, we introduce the MixerCA algorithm, designed for precise classification of hyperspectral imagery across four benchmark datasets. MixerCA employs point-wise convolution to manage channel information and depth-wise convolution to capture spatial details, effectively integrating the benefits of both approaches. Building on insights from previous research, we incorporate large kernel sizes (e.g., $7\times7$) to mix distant spatial locations within each HSI patch, providing a broader receptive field akin to MLPs and self-attention mechanisms. Through extensive testing, MixerCA demonstrated superior performance, particularly for land use and land cover mapping, surpassing the accuracy of other CNN and vision transformer models. For instance, MixerCA achieved an Overall Accuracy of 97.81\% on the Pavia University dataset, exceeding the performance of HybridSN, which reached an OA of 95.67\%. Notably, MixerCA also achieves computational efficiency, requiring 19,145,472 FLOPs and 9,318,144 MACs, in contrast to HybridSN’s 97,483,008 FLOPs and 8,741,504 MACs—highlighting its capacity to deliver high accuracy with reduced resource demands.

Beyond benchmark evaluations, the proposed MixerCA model holds strong potential for real-world applications. Its high accuracy and robustness under limited supervision make it well-suited for critical remote sensing tasks such as land cover mapping, precision agriculture, environmental monitoring, and urban planning. For example, in agriculture, accurate crop-type classification enables optimized irrigation and fertilization strategies. In environmental monitoring, precise land use mapping supports deforestation tracking and wetland conservation. These use cases highlight the broader societal relevance of the proposed approach, extending its utility beyond academic research settings.

\highlighting{Limitations remain in terms of reliance on labeled data, as the model’s performance is influenced by the availability and quality of annotated samples.} Looking forward, our research aims to advance HSI classification in scenarios with limited sample availability through approaches such as self-supervised learning, domain generalization, and semi-supervised learning, thereby reducing dependence on labeled data. Furthermore, techniques like model pruning and knowledge distillation will be explored to optimize model efficiency and reduce computational \highlighting{costs}, while maintaining or improving classification accuracy. \highlighting{Future work should also emphasize scalability to large-scale hyperspectral scenes, requiring efficient strategies such as tiling, hierarchical modeling, and boundary-preserving stitching pipelines. Another promising direction is multimodal integration, where hyperspectral data are fused with complementary sources such as LiDAR, SAR, or high-resolution RGB imagery to improve the discrimination of spectrally similar land-cover types. Finally, enhancing interpretability and trustworthiness is critical, with methods that visualize spectral–spatial features, quantify uncertainty, and provide confidence estimates to support real-world applications in agriculture, urban planning, and environmental monitoring.}

\section*{Acknowledgment}
No funding is available for this work.

\section*{Conflict of Interest}
The authors declare no conflicts of interest.

\printcredits

 \bibliographystyle{model1-num-names}

\bibliography{main}


\end{document}